\documentclass[lettersize,journal]{IEEEtran}
\usepackage{amsmath,amsfonts}
\usepackage{array}
\usepackage[caption=false,font=normalsize,labelfont=sf,textfont=sf]{subfig}
\usepackage{textcomp}
\usepackage{stfloats}
\usepackage{url}
\usepackage{verbatim}
\usepackage{graphicx}
\usepackage{cite}
\usepackage{float}
\usepackage{booktabs}
\usepackage{multirow}
\usepackage[table]{xcolor}
\usepackage{stfloats}
\usepackage{capt-of}
\usepackage{cuted}

\usepackage{xcolor}         
\usepackage{graphicx}
\usepackage{wrapfig}
\usepackage[table,xcdraw]{xcolor}
\usepackage{tabularx}
\usepackage{mathptmx}                  
\usepackage[ruled,vlined,linesnumbered]{algorithm2e}
\usepackage{algpseudocode} 

\title{3D Skew Gaussian Splatting with Any Camera Trajectory Visualization Engine}

\author{Beizhen Zhao, Yifan Zhou, Gaochao Song, Ziran Yin, Hao Wang\textsuperscript{\P}%
\thanks{Beizhen Zhao, Ziran Yin, Hao Wang are with The Hong Kong University of Science and Technology (Guangzhou), Guangzhou, 511442, China
(e-mail:
bzhao610@connect.hkust-gz.edu.cn,
zyin937@connect.hkust-gz.edu.cn,
haowang@hkust-gz.edu.cn),
Yifan Zhou is with Zhejiang University, China
(e-mail:
yifanz@zju.edu.cn),
Gaochao Song is with The University of Hong Kong, China
(e-mail:
gaochaosong\_21@tju.edu.cn).
}
\thanks{\textsuperscript{\P}: Corresponding author.}
}

\begin{document}

\maketitle

\markboth{Journal of \LaTeX\ Class Files,~Vol.~14, No.~8, August~2021}%
{Shell \MakeLowercase{\textit{et al.}}: A Sample Article Using IEEEtran.cls for IEEE Journals}

\begin{abstract}

While 3D Gaussian Splatting (3DGS) has revolutionized real-time photorealistic view synthesis, its fundamental reliance on symmetric Gaussian distributions introduces visual artifacts that hinder accurate spatial data exploration. 
Specifically, symmetric kernels struggle to capture shape and color discontinuities
, which cause blurriness and primitive redundancy that mislead human perception during visual analysis. 
To address these visualization barriers, we introduce 3D Skew Gaussian Splatting (3DSGS), a novel framework that significantly enhances the structural fidelity and compactness of explicit scene representations. 
Our key insight lies in extending the standard primitive to a general Skew Gaussian counterpart. 
This generalized primitive inherits the highly efficient rasterization properties of standard Gaussians while gaining intrinsic asymmetric modeling capabilities. 
We couple this with an enhanced opacity representation to better handle complex transparency, alongside a depth-aware densification strategy that intelligently manages primitive allocation. 
Furthermore, to make these advancements actionable for real-world visual analytics, we re-derive the CUDA rasterization pipeline to universally support both symmetric and skew Gaussians, integrating it into a decoupled, free-camera interactive visualization engine. 
Extensive experiments demonstrate that 3DSGS achieves superior rendering quality and structural compactness, particularly in regions with intricate details, while maintaining the real-time frame rates necessary for fluid interactive exploration. Supplementary derivations and visual results are available at \textbf{\textit{https://3d-skew-gs.github.io/}}.
\end{abstract}


\section{Introduction}

\begin{figure*}
  \centerline{\includegraphics[width=1.0\textwidth]{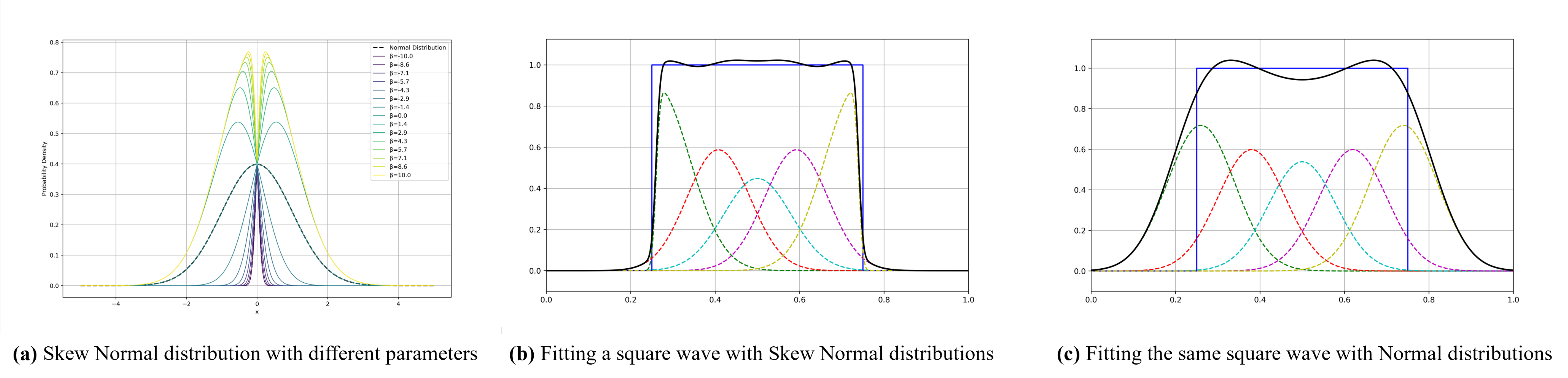}}
  \caption{Comparison of Skew Normal and Normal Kernels in Fitting a Square Wave.
(a): Distribution shapes with different skew parameters in the Skew Normal distribution, where the Gaussian distribution is the special case when the skewness coefficient is zero.
(b): Fitting a square wave using five Skew Normal distributions, demonstrating a closer approximation of sharp edges due to adjustable skewness. 
(c): Fitting the same square wave with five standard Normal distributions, which produce smoother approximations and less accurate edge representation. 
  }
  \label{s1}
\end{figure*}

Reconstructing and visually exploring realistic 3D scenes from multi-view images is a cornerstone challenge in computer graphics and spatial data analysis. Recent advancements have heavily favored explicit or hybrid representations to achieve the interactive rendering speeds required for real-time visual analytics~\cite{choi2022time, wen2023bundlesdf, tang2024sparseocc, wang2025look, zhai2025splatloc}. Among these, 3D Gaussian Splatting (3DGS)~\cite{kerbl20233d} has emerged as a state-of-the-art technique. By representing scenes as a collection of 3D Gaussians and utilizing an efficient rasterization pipeline, 3DGS enables photorealistic, real-time rendering. 
This capability is paramount not only for automated systems but also for human-driven visual exploration, allowing analysts to interact dynamically with complex spatial environments~\cite{kheradmand20243d, kulhanek2024wildgaussians, zha2024r}.

Despite its remarkable success in view synthesis, standard 3DGS introduces visual artifacts that hinder accurate data interpretation. A symmetric Gaussian kernel, defined solely by its mean and covariance, intrinsically struggles to represent asymmetric spatial features. 
Real--world phenomena, such as specular reflections, view-dependent shading along sharp geometric boundaries, and complex transparency, often exhibit strong directional asymmetry~\cite{jiang2024gaussianshader, ye20243d, yang2024spec, zakharov2024human}. 
To approximate these features, standard 3DGS forces the deployment of densely overlapping or excessively diminutive primitives. This work-around frequently introduces blurriness, visual discontinuities, and bloated data footprints that compromise both rendering performance and visual fidelity~\cite{chen2024beyondgaussian, rota2024revising}. 
Furthermore, the standard opacity model struggles to flexibly render sharp opacity transitions, further obfuscating the accurate perception of occlusions and transparent objects~\cite{xu2024supergaussians, hyung2024effective}.

To address these visualization barriers, we propose 3D Skew Gaussian Splatting (3DSGS), a framework that introduces a more expressive asymmetric kernel to accurately capture spatial discontinuities. 
As illustrated in Fig. \ref{s1}(b) and (c), fitting a square wave with five skew gaussian distributions yields a better representation than using five gaussian distributions, as the skew parameter allows for sharper edges.
This improvement can be attributed to the ability of the Skew Normal distribution to capture higher frequency content more effectively than the standard Gaussian and as shown in Fig.~\ref{s1}(a), the gaussian distribution is the special case of the skew gaussian distribution when the skewness is zero.
By combining this enhanced primitive with a depth-aware densification strategy and an extended opacity model, 3DSGS provides a more flexible scene representation that is better suited for the challenges of real-world robot vision and spatial perception, while retaining the crucial real-time rendering performance.

To realize this, we systematically rederive and implement the rasterization rendering pipeline from the ground up using CUDA. This highly optimized, custom rasterizer is designed for universal compatibility, seamlessly adapting to both standard 3DGS and our proposed 3DSGS without compromising interactive frame rates. 
Crucially, to support the complex visual analysis of these scenes, we integrate this generalized CUDA backend into a dedicated visualization engine within Blender. 
This integration empowers users with unconstrained, free camera movement, allowing them to seamlessly navigate, render, and visually explore the optimized clusters in real-time. 
By bridging expressive asymmetric primitives with robust underlying algorithms and interactive tooling, our framework ensures that analysts can reliably perceive and interact with high-fidelity spatial data.

Our primary contributions are fourfold:

\begin{itemize}
    \item \textbf{3D Skew Gaussian Primitives:} We fundamentally advance the core 3DGS representation by incorporating a 3D Skew Normal distribution. This introduces a skewness parameter that captures asymmetric, view-dependent appearances, significantly improving the visual fidelity of sharp boundaries and reflections.
    \item \textbf{Enhanced Opacity and Densification:} We develop an extended opacity representation to model complex transparency alongside a novel depth-aware densification strategy. This intelligent allocation prunes redundant splats, resulting in a more compact and visually accurate geometric representation.
    \item \textbf{Generalized CUDA Rasterization Pipeline:} We completely re-derive and implement an optimized rasterization pipeline. This robust CUDA backend acts as a universal engine, seamlessly accommodating both standard and skew Gaussians, maximizing rendering efficiency and algorithmic generality.
    \item \textbf{Interactive Visualization Engine:} We design and integrate a visualization framework supporting free camera movement within Blender, providing a powerful interactive environment for users to freely explore, render, and analyze 3DSGS scenes.
\end{itemize}

\begin{figure*}
  \centerline{\includegraphics[width=1.0\textwidth]{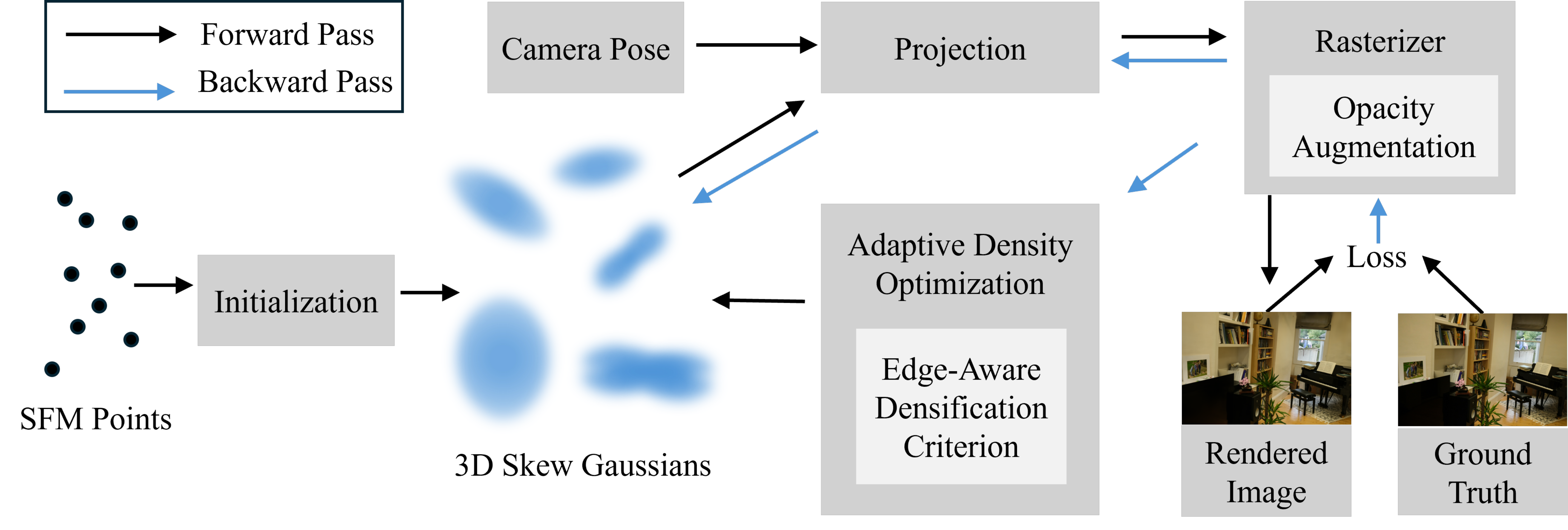}}
  \caption{The architectural pipeline of our 3D Skew Gaussian Splatting (3DSGS) framework. 
  We fundamentally extend standard symmetric 3D Gaussians into 3D Skew Gaussians by introducing an intrinsic skewness parameter, empowering the model to capture complex spatial discontinuities and asymmetric structures. 
  We systematically re-derive and implement a generalized splatting pipeline that seamlessly handles both the forward rendering and backward optimization passes for asymmetric kernels. 
  During the training stage, a novel depth-aware densification criterion intelligently manages the splitting and pruning of Skew Gaussians, ensuring high structural compactness. 
  Ultimately, the optimized scene representation is decoupled and exported to our VisEngine for real-time, interactive visual exploration.
  }
  \label{fig:framework}
\end{figure*}

\section{Related Work}

\subsection{Novel View Synthesis and Interactive Exploration}

Early approaches to Novel View Synthesis (NVS) and spatial data visualization frequently relied on explicit geometry reconstruction, such as mesh or point cloud-based representations. While these methods offered intuitive ways to inspect 3D environments, they often struggled to render complex scenes and view-dependent lighting effects with high fidelity. Recent breakthroughs have largely been driven by neural and volumetric rendering techniques, shifting the paradigm toward highly realistic, data-driven scene representations.

\noindent\textbf{Implicit Neural Scene Representations:} 
Implicit methods, most notably Neural Radiance Fields (NeRF)~\cite{mildenhall2021nerf}, pioneered the use of Multilayer Perceptrons (MLPs) to encode continuous density and radiance fields, achieving photorealistic novel views through volumetric ray marching. Subsequent works, such as Mip-NeRF~\cite{barron2021mip} and Instant-NGP~\cite{muller2022instant}, significantly improved training efficiency and spatial consistency using positional encoding, hierarchical sampling, and factorized grid architectures. These NeRF-based models excel at capturing intricate geometry and complex view-dependent appearances. However, for the visualization community, the computationally intensive nature of ray marching and continuous network evaluations imposes a severe bottleneck, restricting the interactive frame rates required for fluid, user-driven spatial exploration.

\noindent\textbf{Explicit Point-Based and Gaussian Representations:}
Before the advent of 3DGS, explicit point-based rendering methods (e.g., ADOP~\cite{ruckert2022adop}) demonstrated the potential of combining point clouds with neural deferred shading for faster rendering. Building on this philosophy to bridge the gap between photorealistic synthesis and real-time interactivity, 3D Gaussian Splatting (3DGS)~\cite{kerbl20233d} has emerged as a transformative technique~\cite{yang2024deformable}. By modeling scenes as explicit 3D Gaussian primitives, 3DGS leverages spatial locality, transparency modeling, and highly efficient tile-based rasterization~\cite{chung2023luciddreamer, xu2024wild, zhou2024dreamscene360}. This explicit structure makes it exceptionally well-suited for integration into interactive visual engines. However, its strict reliance on standard, symmetric Gaussian kernels limits its expressiveness when modeling complex spatial structures. Phenomena such as sharp geometric boundaries, intricate transparency, and strong directional view-dependent effects often force the optimization process to over-compensate by deploying massive numbers of diminutive primitives~\cite{cao20243dgs, sun2024f3dgs, morgenstern2024compact}. In interactive visualization, this structural redundancy inflates the memory footprint and introduces visual artifacts (e.g., blurring and popping) that can mislead user perception during visual analysis.

\subsection{Enhanced Gaussian Kernels and Structural Fidelity}

To mitigate the limitations of standard symmetric kernels, a recent wave of studies has focused on improving the expressiveness, structural alignment, and compactness of Gaussian primitives. Efforts to reduce visual artifacts have primarily explored modifying the basic shape or geometric constraints of the primitives. Student Splatting and Scooping~\cite{zhu20253d} replaces the standard kernel with a learnable Student's t-distribution, enabling more flexible kernel shapes and improved geometric adaptation. To address the volume-bloating issue and better approximate thin structures, 2D Gaussian Splatting (2DGS)~\cite{huang20242dgs} collapses the 3D volume into oriented 2D disks, significantly improving surface extraction and perspective-accurate rendering. To further enhance visual fidelity around sharp edges, 3D Half-Gaussian Splatting~\cite{li20243d} introduces half-Gaussian kernels with distinct opacity values. In a similar vein, DRK Splatting~\cite{huang2024deformable} proposes a learnable non-linear edge-sharpening function to improve rendering quality on view-dependent surfaces and complex occlusions. 

Beyond isolated shape modifications, managing multi-scale details and suppressing rendering artifacts in structurally complex scenes has led to frequency-domain and multiresolution enhancements. For instance, Mip-Splatting~\cite{yu2024mip} introduced a 3D smoothing filter to eliminate high-frequency aliasing artifacts during camera zooming. 
Extending this logic to frequency-domain scene representations, Wavelet-GS~\cite{waveletgs2025} employs wavelet decomposition to systematically manage multi-scale geometric features, enhancing visual fidelity and suppressing noise without causing an explosion in primitive count. Furthermore, addressing the entanglement of geometry and texture, methods like 3D Convex Splatting~\cite{held20243d} utilize smooth convexes to decouple geometric features from appearance, allowing for more efficient optimization and cleaner structural boundaries.

These works share a common goal: pushing the boundaries of how explicit primitives represent geometric and appearance complexity to eliminate visual artifacts. 
However, a fundamental limitation remains that most still operate under the assumption of symmetric or isotropic core kernels. 
Consequently, they must rely on indirect mechanisms to capture spatial asymmetry. 
In an interactive visualization context, these workarounds not only complicate the optimization landscape but also introduce significant computational overhead during the forward rasterization pass, limiting fluid exploration. 
In contrast, we address this challenge by explicitly introducing the 3D Skew Normal distribution as the core primitive. By empowering each splat with an intrinsic asymmetry parameter, our method directly captures directional phenomena at the mathematical foundation of the kernel. 
This direct approach requires significantly fewer primitives to achieve high structural compactness and ultimately ensures the robust performance required for real-time interactive visual analytics.

\subsection{Rendering Pipelines and Visualization Frameworks}

The transition from static novel view synthesis to dynamic visual exploration relies heavily on robust underlying rendering pipelines. The original 3DGS implementation features a highly specialized CUDA rasterizer optimized exclusively for standard covariance matrices. When attempting to introduce complex primitives—such as asymmetric distributions or customized projection mechanisms—researchers often face significant performance degradation or are forced to rely on approximate blending workarounds. Consequently, developing flexible, generalized rasterization architectures has become a pressing need.

Moreover, making these advanced scene representations accessible for practical spatial analysis requires seamless integration into established 3D authoring and exploration tools. While standalone viewers exist, integrating customized splatting backends directly into comprehensive suites like Blender empowers users with unconstrained, free-camera movement and deep analytical capabilities. Our work directly addresses this gap by re-deriving and implementing a generalized CUDA rasterizer that natively supports both standard symmetric and novel skew Gaussians. This robust backend serves as the high-performance foundation for our interactive Blender visualization engine, ensuring that advanced representation techniques are fully actionable for user-driven visual exploration.

\section{Preliminaries}
\label{sec:preliminaries}

\subsection{Preliminary of 3D Gaussian Splatting}

Formally, 3DGS represents a scene as a collection of $N$ anisotropic 3D Gaussians $\{G_i\}_{i=1}^N$. Each primitive is parameterized by a continuous 3D probability density function:
\begin{equation}
    G_i(\mathbf{x}) = \exp\left(-\frac{1}{2}(\mathbf{x}-\boldsymbol{\mu}_i)^T \boldsymbol{\Sigma}_i^{-1} (\mathbf{x}-\boldsymbol{\mu}_i)\right),
\end{equation}
where $\boldsymbol{\mu}_i \in \mathbb{R}^3$ denotes the spatial center (mean) of the primitive, and $\boldsymbol{\Sigma}_i \in \mathbb{R}^{3\times3}$ is the 3D covariance matrix determining its scale and orientation. 
To ensure that $\boldsymbol{\Sigma}_i$ remains physically valid during gradient-based optimization, it is typically factorized into a learnable scaling matrix $\mathbf{S}_i$ and a rotation matrix $\mathbf{R}_i$ (derived from a quaternion), such that $\boldsymbol{\Sigma}_i = \mathbf{R}_i \mathbf{S}_i \mathbf{S}_i^T \mathbf{R}_i^T$. Each Gaussian is additionally coupled with a base opacity $\alpha_i$ and view-dependent color coefficients parameterized via Spherical Harmonics (SH).

To synthesize a novel view, these 3D primitives are projected onto the 2D image plane. This splatting process employs an affine approximation of the viewing transformation to compute a 2D covariance matrix $\boldsymbol{\Sigma}_i'$, which defines the elliptical footprint of the projected Gaussian on the screen. 
The scene is then rendered pixel by pixel utilizing a highly efficient, tile-based rasterization pipeline. For a given pixel, the intersecting 2D Gaussians are sorted in depth, and their color contributions are accumulated using front-to-back alpha blending:
\begin{equation}
    C = \sum_{i} c_i \alpha_i' \prod_{j=1}^{i-1} (1-\alpha_j'),
\end{equation}
where $c_i$ is the view-dependent color evaluated from the SH coefficients, and $\alpha_i'$ is the final opacity of the $i$-th Gaussian, computed as the product of the learned base opacity $\alpha_i$ and the 2D Gaussian falloff. 
Because this entire model is analytically differentiable, the attributes of all Gaussians can be optimized by gradient descent, driven by photometric loss formulations between the rendered outputs and ground-truth images.

\subsection{Interactive 3DGS Exploration Workflow in Blender}
\label{sec:blender_workflow}

To provide analysts with an intuitive and unconstrained environment for spatial data exploration, VisEngine leverages Blender as its interactive frontend. 
The operational workflow is designed to bridge the gap between abstract mathematical representations and human-centric visual analysis without requiring users to write visualization scripts. 

The workflow begins by importing a visual proxy of the 3DSGS scene into the Blender viewport. 
Depending on the analytical requirement, this proxy can either be the raw, colored point cloud extracted from the initialized Gaussians (imported via our customized \texttt{.ply} parser), or a continuous geometric mesh (e.g., an \texttt{.obj} file generated via surface reconstruction). 
These proxies provide immediate spatial context. 
Analysts then utilize Blender's native viewport controls to perform unconstrained free camera movements. 
They can fluidly orbit around complex spatial structures, inspect asymmetric specular reflections from specific angles, or establish keyframes along a timeline to design precise, cinematic analytical trajectories.
Because the heavy CUDA rasterization is completely decoupled from this frontend interaction loop, analysts experience zero latency during trajectory design, ensuring a seamless and responsive exploratory data analysis experience.

\subsection{Camera Trajectories and Format Standardization}
\label{sec:camera_formats}

A significant bottleneck in standardizing interactive radiance fields is the fragmentation of camera coordinate conventions across different computational domains. 
Synthesizing novel views relies on perfect synchronization between the frontend trajectory design and the backend CUDA rasterizer. 
However, these two ecosystems fundamentally disagree on the definition of local camera space.

\noindent\textbf{The OpenCV / COLMAP Convention:}
In the computer vision ecosystem, most Structure-from-Motion (SfM) algorithms such as COLMAP and explicit neural rendering backends operate under the OpenCV camera coordinate standard. 
In this convention, the camera looks down the positive Z-axis ($+Z$), the local Y-axis points downwards ($+Y$ down), and the X-axis points to the right ($+X$ right). 
Camera extrinsic parameters outputted by these pipelines typically store Camera-to-World (C2W) matrices strictly adhering to this Right-Down-Forward structural assumption.

\noindent\textbf{The OpenGL / Blender Convention:}
In contrast, interactive 3D authoring environments and graphics APIs heavily rely on the OpenGL standard. 
When an analyst navigates the free camera within Blender, the generated local camera coordinate system assumes the camera looks down the negative Z-axis ($-Z$), the local Y-axis points upwards ($+Y$ up), and the X-axis points to the right ($+X$ right). 
Consequently, directly feeding raw Blender camera parameters into a 3DGS rasterizer results in severe axis-flipping artifacts, rendering the scene upside down or completely out of the field of view.

\section{Methodology}

3D Gaussian Splatting (3DGS) models spatial environments using symmetric volumetric kernels. 
However, these symmetric primitives inherently struggle to accurately represent spatial discontinuities. 
To overcome these representational barriers, we propose 3D Skew Gaussian Splatting (3DSGS). 
By integrating the Skew Normal distribution into the splatting architecture, we introduce a more robust rendering pipeline, naturally encompassing the standard symmetric Gaussian as a special baseline case. 

The remainder of this section details our systematic approach. Section 3.1 briefly reviews the mathematical foundations of classical 3DGS. In Section 3.2, we formulate the 3D Skew Gaussian kernel and its asymmetric properties. Section 3.3 details our rigorous extensions to both the forward and backward passes of the rasterization pipeline. Section 3.4 outlines a novel depth-aware densification criterion designed to optimize primitive allocation and ensure structural compactness. Finally, Section 3.5 describes the integration of these core rendering components into a decoupled interactive visualization engine (VisEngine) to facilitate unconstrained visual analysis. The comprehensive training and rendering workflow of our 3DSGS framework is illustrated in Figure~\ref{fig:framework}.

\begin{algorithm}[h]
\caption{Forward Pass of 3DSGS}
\label{alg:forward}
\KwIn{Pixel coordinate $\mathbf{x}$, Gaussian mean $\boldsymbol{\mu}$, inverse covariance parameters $a, b, c$, skewness coefficients $\beta_x, \beta_y$, base opacity $d$, color features $c_i$}
\KwOut{Rendered pixel color $C$}
\BlankLine
\tcp{Compute spatial offsets}
$\delta_x \leftarrow x - \mu_x$, $\delta_y \leftarrow y - \mu_y$\;
\BlankLine
\tcp{Compute standard 2D Gaussian and skewness terms}
$G^{\prime}(\mathbf{x}) \leftarrow \exp\left(-\frac{1}{2}(a\delta_x^2 + c\delta_y^2) - b\delta_x\delta_y\right)$\;
$z \leftarrow \frac{\beta_x\delta_x + \beta_y\delta_y}{\sqrt{2}}$\;
$\Phi(\mathbf{x}) \leftarrow 1 + \operatorname{erf}(z)$\;
\BlankLine
\tcp{Compute final modulated opacity}
$\alpha \leftarrow d \cdot G^{\prime}(\mathbf{x}) \cdot \Phi(\mathbf{x})$\;
\BlankLine
\tcp{Alpha blending for volume rendering}
$T_1 \leftarrow 1$\;
$C \leftarrow 0$\;
\For{each intersected Gaussian $i$ from near to far}{
    $C \leftarrow C + c_i \cdot \alpha_i \cdot T_i$\;
    $T_{i+1} \leftarrow T_i \cdot (1 - \alpha_i)$\;
}
\Return $C$\;
\end{algorithm}

\subsection{3D Skew Gaussian Kernel}
Traditional 3DGS models scene components using symmetric Gaussian kernels, characterized solely by their mean \(\boldsymbol{\mu}\) and covariance \(\boldsymbol{\Sigma}\). While effective in representing isotropic and symmetric features, such models are limited in their ability to accurately depict asymmetric or discontinuous structures commonly found at object edges, corners, or textured regions. To overcome these limitations, we propose a skew Gaussian kernel formulated as:

\begin{equation}
S(\mathbf{x}) = 2 \, G(\mathbf{x}; \boldsymbol{\mu}, \boldsymbol{\Sigma}) \Phi\left( {\beta}^T (\mathbf{x} - \boldsymbol{\mu}) \right),
\end{equation}

where \(G(\mathbf{x}; \boldsymbol{\mu}, \boldsymbol{\Sigma})\) is the conventional multivariate Gaussian density, and \(\Phi(\cdot)\) denotes the cumulative distribution function (CDF) of the standard normal distribution. \(\boldsymbol{\beta} \in \mathbb{R}^d\) denotes Skewness vector controlling the direction and degree of asymmetry along each axis. 
Larger magnitudes indicate stronger skewness, allowing the kernel to stretch or compress asymmetrically.

The derivative of $S$ with respect to $\beta$ is given by the formula in Eq. \ref{eq:sbeta}. 
The detailed derivation process can be found in the supplementary materials.

\begin{equation}
\frac{\partial S(\mathbf{x})}{\partial \boldsymbol{\beta}} = G(\mathbf{x}; \boldsymbol{\mu}, \boldsymbol{\Sigma}) \cdot \frac{2}{\sqrt{\pi}} \cdot e^{ - \left( \frac{\boldsymbol{\beta}^T (\mathbf{x} - \boldsymbol{\mu})}{\sqrt{2}} \right)^2 } \cdot \frac{1}{\sqrt{2}} (\mathbf{x} - \boldsymbol{\mu}).
\label{eq:sbeta}
\end{equation}

This enhanced kernel design allows for more flexible and precise scene representations, particularly beneficial for modeling real-world scenes with complex structures.
By accommodating directional biases in density distribution, the model can achieve more faithful reconstructions, especially in regions with abrupt changes. 
Additionally, the skew-normal formulation preserves smoothness and compatibility with gradient-based optimization.

The opacity value alpha, used in the process of alpha blending is determined through a specific calculation to accurately represent the transparency level of each element:

\begin{align}
\alpha
=o_i\cdot S_i(x) = o_i\cdot G_i(x) \cdot \Phi\left( {\beta}^T  ({x} - {\mu}) \right),\\
\label{eq:alpha}
\end{align}
where

\begin{equation}
\Phi(z) = \frac{1}{2} \left[ 1 + \operatorname{erf}\left(\frac{z}{\sqrt{2}}\right) \right].
\end{equation}

The 3D Gaussian distribution utilized in 3DGS is precisely the special case of the Skew Normal distribution where the skewness vector $\boldsymbol{\beta} = \mathbf{0}$. 
In this case, we have $\Phi(0) = 0.5$, then the Skew Normal PDF simplifies to:
\begin{equation}
    S(\mathbf{x}; \boldsymbol{\mu}, \boldsymbol{\Sigma}, \mathbf{0}) = 2 G(\mathbf{x}; \boldsymbol{\mu}, \boldsymbol{\Sigma}) \Phi(0) = G(\mathbf{x}; \boldsymbol{\mu}, \boldsymbol{\Sigma}).
\end{equation} 

Our Skew Normal primitive thus generalizes the 3DGS primitive, allowing for a continuous transition from symmetric to arbitrarily skewed distributions.
The power of the Skew Normal kernel lies in the $\Phi\left( \boldsymbol{\beta}^T (\mathbf{x} - \boldsymbol{\mu}) \right)$ term, which acts as a directional modulation factor on the underlying symmetric Gaussian $G(\mathbf{x})$. The scalar value $\boldsymbol{\beta}^T (\mathbf{x} - \boldsymbol{\mu})$ represents the projection of the vector $(\mathbf{x} - \boldsymbol{\mu})$ onto the direction of the skewness vector $\boldsymbol{\beta}$. The sign of this projection determines whether $\mathbf{x}$ is on the "positive" or "negative" side of the hyperplane passing through $\boldsymbol{\mu}$ with normal $\boldsymbol{\beta}$. The CDF $\Phi(\cdot)$ then maps this scalar projection to a value between 0 and 1.

\begin{algorithm}[h]
\caption{Backward Pass of 3DSGS}
\label{alg:backward}
\KwIn{Loss gradient w.r.t color $\frac{\partial L}{\partial C}$, forward pass intermediate variables $\delta_x, \delta_y, z, G^{\prime}(\mathbf{x}), S^{\prime}(\mathbf{x}), \alpha$}
\KwOut{Gradients for learning parameters: $\frac{\partial L}{\partial x_{ndc}}, \frac{\partial L}{\partial y_{ndc}}, \frac{\partial L}{\partial a}, \frac{\partial L}{\partial b}, \frac{\partial L}{\partial c}, \frac{\partial L}{\partial \beta_x}, \frac{\partial L}{\partial \beta_y}$}
\BlankLine
\tcp{Compute base gradient for opacity $\alpha_k$}
$\frac{\partial L}{\partial \alpha_k} \leftarrow T_k \sum_i \frac{\partial L}{\partial C[i]}(c_k[i] - R_{k+1})$\;
\BlankLine
\tcp{Gradients for screen-space coordinates}
$\frac{\partial L}{\partial x_{ndc}} \leftarrow -d \cdot \frac{\partial L}{\partial \alpha_k} \cdot G^{\prime}(\mathbf{x}) \cdot \frac{W}{2} \left[ (a\delta_x + b\delta_y)(1 + \operatorname{erf}(z)) + \sqrt{\frac{2}{\pi}} \beta_x \exp(-z^2) \right]$\;
$\frac{\partial L}{\partial y_{ndc}} \leftarrow -d \cdot \frac{\partial L}{\partial \alpha_k} \cdot G^{\prime}(\mathbf{x}) \cdot \frac{H}{2} \left[ (b\delta_x + c\delta_y)(1 + \operatorname{erf}(z)) + \sqrt{\frac{2}{\pi}} \beta_y \exp(-z^2) \right]$\;
\BlankLine
\tcp{Gradients for inverse covariance parameters}
$\frac{\partial L}{\partial a} \leftarrow -\frac{1}{2} d \cdot \delta_x^2 \cdot S^{\prime}(\mathbf{x}) \frac{\partial L}{\partial \alpha_k}$\;
$\frac{\partial L}{\partial b} \leftarrow -d \cdot \delta_x \delta_y \cdot S^{\prime}(\mathbf{x}) \frac{\partial L}{\partial \alpha_k}$\;
$\frac{\partial L}{\partial c} \leftarrow -\frac{1}{2} d \cdot \delta_y^2 \cdot S^{\prime}(\mathbf{x}) \frac{\partial L}{\partial \alpha_k}$\;
\BlankLine
\tcp{Gradients for skewness coefficients}
$E_{skew} \leftarrow \exp\left(-\frac{1}{2}(a\delta_x^2 + c\delta_y^2 + 2z^2) - b\delta_x\delta_y\right)$\;
$\frac{\partial L}{\partial \beta_x} \leftarrow d \cdot \frac{\partial L}{\partial \alpha_k} \sqrt{\frac{2}{\pi}} \delta_x \cdot E_{skew}$\;
$\frac{\partial L}{\partial \beta_y} \leftarrow d \cdot \frac{\partial L}{\partial \alpha_k} \sqrt{\frac{2}{\pi}} \delta_y \cdot E_{skew}$\;
\BlankLine
\Return All computed gradients\;
\end{algorithm}

\subsection{3D Skew Gaussian Rasterization and Splatting}
\label{sec:rasterization}

Building upon the highly efficient tile-based architecture of standard 3DGS, we extend the rasterization and splatting pipeline to accommodate the geometric complexity of Skew Gaussian kernels. 
During the forward pass, the 2D image is rendered by accumulating the contributions of each projected skew Gaussian through alpha blending. 
The final color $C$ of a given pixel is computed by evaluating the intersecting primitives in a strictly back-to-front depth order:
\begin{equation}
C = \sum_{i\in\mathcal{N}} c_i \alpha_i(\hat{\mathbf{x}}-\hat{\boldsymbol{\mu}}) \prod_{j=1}^{i-1} \left(1-\alpha_j(\hat{\mathbf{x}}-\hat{\boldsymbol{\mu}})\right).
\label{eq:color}
\end{equation}

Unlike standard symmetric Gaussians that utilize a static scalar opacity, our 3DSGS framework introduces a spatially varying opacity model to capture complex transparency and sharp occlusion boundaries.
Inspired by recent advancements in structural modeling, we incorporate an additional analytical term that modulates the transparency across the internal footprint of a single kernel. The effective opacity $o$ evaluated at the 2D projected coordinate $\mathbf{x}$ is defined as:
\begin{equation}
    o(\mathbf{x}) = \frac{1}{2}\left\{ (\alpha_1+\alpha_2) + (\alpha_1-\alpha_2)\operatorname{erf}\left(\frac{(\boldsymbol{\beta}+\mathbf{d})^T \mathbf{x}}{\sqrt{2}}\right) \right\},
\label{eq:alpha}
\end{equation}
where $\alpha_1$ and $\alpha_2$ denote the distinct base opacities of the bisected regions, $\boldsymbol{\beta}$ is the intrinsic skewness parameter that governs the asymmetry of the distribution, and $\mathbf{d}$ is a directional vector guiding the internal segmentation boundary. This formulation empowers a single skew Gaussian to flexibly represent sharp opacity transitions and directional variations, significantly reducing the visual blurring typically caused by the forced overlapping of multiple symmetric primitives.

Because this customized splatting formulation is fully analytically differentiable, we seamlessly integrate it into the backward pass for end-to-end optimization. Driven by the photometric loss $\mathcal{L}$ between the synthesized and ground-truth images, the gradients must propagate through our complex opacity formulation. Specifically, the gradient with respect to the skewness parameter $\boldsymbol{\beta}$ is derived via the chain rule:
\begin{equation}
\frac{\partial \mathcal{L}}{\partial \boldsymbol{\beta}} = \frac{\partial \mathcal{L}}{\partial \alpha_i} \cdot \frac{\partial \alpha_i}{\partial \mathbf{S}} \cdot \frac{\partial \mathbf{S}}{\partial \boldsymbol{\beta}},
\label{eq:lbeta}
\end{equation}
where $\mathbf{S}$ represents the intermediate state variables of the kernel footprint. 
This gradient flow ensures that the model penalizes structural discrepancies, promoting stable convergence. 
Detailed mathematical derivations are provided in the supplementary material.

\subsection{Depth-Aware Densification Strategy}
\label{sec:densification}

A critical component of explicit radiance fields is the adaptive optimization of the point cloud density. 
Standard 3DGS heavily relies on the 2D screen-space positional gradient norm, denoted as $g_i^{(uv)}$, to trigger this densification. However, in the context of high-fidelity visual analytics, we observe that relying solely on view-dependent 2D projected errors is insufficient to capture complex topologies. Specifically, regions near sharp occlusion boundaries and depth discontinuities often suffer from under-reconstruction or floaters, which severely degrade human depth perception during interactive 3D exploration.

To enforce geometric fidelity strictly along the viewing ray and eliminate these visual artifacts, we introduce a depth-aware densification constraint. We quantify the sensitivity of each primitive to depth perturbations. For each skew Gaussian indexed by $i$, we define the maximal gradient magnitude with respect to the depth coordinate over the optimization history as:
\begin{equation}
g_i^{(z)} = \max_t \left\| \nabla_z \mathcal{L}_i^t \right\|_2,
\label{eq:depth_grad}
\end{equation}
where $\mathcal{L}_i^t$ represents the rendering loss for Gaussian $i$ at training iteration $t$, and $\nabla_z \mathcal{L}_i^t$ denotes the explicit gradient of this loss with respect to the depth ($Z$-axis) of the primitive's spatial center. 

By incorporating $g_i^{(z)}$ into the adaptive control loop alongside the standard 2D gradient $g_i^{(uv)}$, our strategy intelligently identifies regions with severe depth ambiguities. Gaussians exhibiting high depth-gradients are prioritized for splitting or cloning. 
This depth-aware mechanism ensures that skew Gaussians are distributed compactly and accurately along physical surface boundaries, resulting in a more reliable geometric representation for spatial analysis.

\begin{figure}
  \centerline{\includegraphics[width=0.48\textwidth]{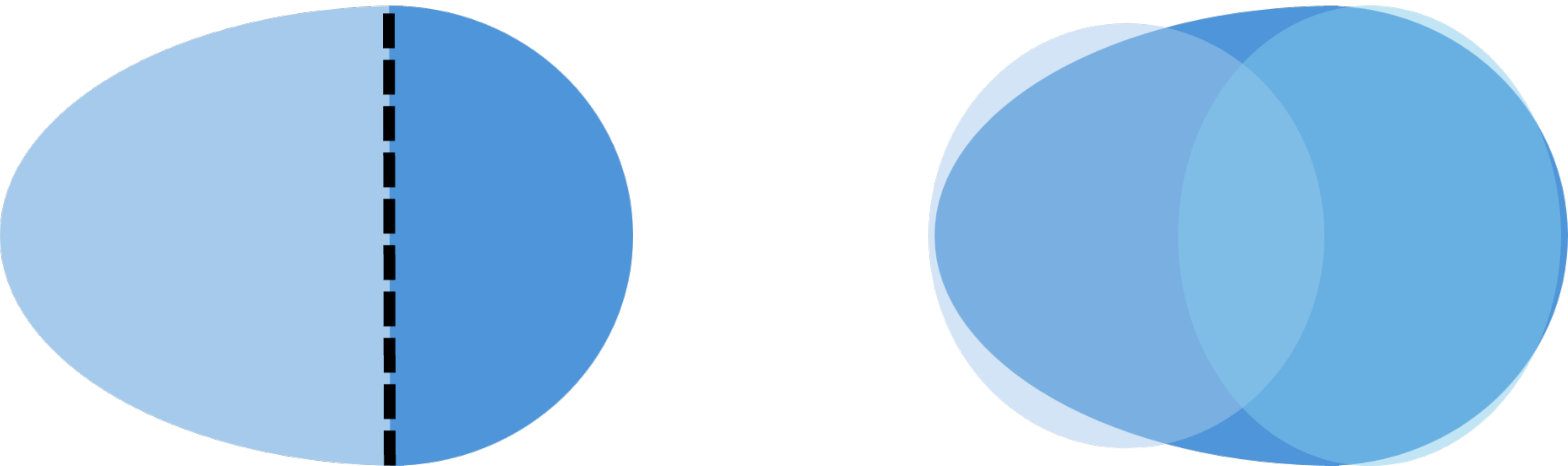}}
  \caption{The left sub-figure shows how the enhanced opacity parameter affects the skew kernel. The right sub-figure indicates split the skew kernel along the max scale direction and adjust the scale of new kernels.}
  \label{split}
\end{figure}

Then we improve the split strategy during the densification process.
We split the skew kernel on the max axis and adjust the scale of new kernels after splitting with the skew parameters to fit the original shape as shown in Fig.~\ref{split}.
This split strategy can help us generate more accurate Gaussians.

\begin{algorithm}[tb]
\caption{Decoupled Interactive Rendering Pipeline for 3DSGS}
\label{alg:visengine}

\KwIn{Optimized 3DSGS primitive set $\mathcal{P}$, viewport resolution $(W,H)$}
\KwOut{Real-time photorealistic novel view $\mathcal{I}_t$ at frame $t$}

$\mathbf{T}_{align} \leftarrow \mathrm{diag}(1,-1,-1,1)$ \tcp{OpenGL $\rightarrow$ OpenCV transform}

\While{interactive visual exploration}{

\textbf{Phase 1: Spatial Context \& Trajectory Acquisition}

$\mathbf{M}_{gl}^{(t)} \leftarrow$ frontend camera pose (C2W) \;

$\mathbf{K}^{(t)} \leftarrow$ intrinsic matrix derived from $(W,H)$ \;

\vspace{2pt}

\textbf{Phase 2: Coordinate Alignment (CoordEngine)}

$\mathbf{M}_{cv}^{(t)} \leftarrow \mathbf{M}_{gl}^{(t)} \mathbf{T}_{align}$ \tcp{coordinate consistency}

\vspace{2pt}

\textbf{Phase 3: Backend Synthesis \& Visual Feedback}

$\mathcal{I}_t \leftarrow \text{CUDARasterizer}(\mathcal{P},\mathbf{M}_{cv}^{(t)},\mathbf{K}^{(t)})$ \;

UpdateViewport$(\mathcal{I}_t)$ \;

}

\end{algorithm}

\subsection{Decoupled Interactive Visual Analytics Engine}
\label{sec:visengine}

A challenge in applying advanced radiance fields to visual analytics is the tight coupling between the rendering pipelines and restricted viewing interfaces. 
This rigidity severely hinders the exploratory data analysis (EDA) required to understand complex spatial phenomena. To bridge the gap between high-fidelity view synthesis and human-centric spatial cognition, we propose a decoupled Interactive Visual Analytics Engine (VisEngine). 
By integrating our generalized CUDA rasterization backend into a mature 3D authoring environment (Blender), our method shifts the paradigm from passive viewing to active, unconstrained visual exploration. The architectural flow is formalized in Algorithm~\ref{alg:visengine}.

Effective spatial analysis relies heavily on accurate contextual awareness. 
A primary obstacle when integrating computer vision reconstruction techniques with standard visual design platforms is the fundamental mismatch in coordinate systems. 
Our rendering pipeline, derived from Structure-from-Motion (SfM) conventions (e.g., COLMAP), operates in the OpenCV coordinate space. 
Conversely, interactive environments like Blender utilize the OpenGL standard (Right-Up-Back). To prevent spatial disorientation and axis-flipping artifacts during visual analysis, VisEngine introduces an embedded \textit{CoordEngine}. For any viewpoint $\mathbf{M}_{\text{gl}}$ interactively defined by the user, CoordEngine mathematically guarantees alignment through a transformation matrix $\mathbf{T}_{\text{align}}$:
\begin{equation}
    \mathbf{M}_{\text{cv}} = \mathbf{M}_{\text{gl}} \mathbf{T}_{\text{align}} = \mathbf{M}_{\text{gl}} 
    \begin{bmatrix} 
        1 &  0 &  0 & 0 \\ 
        0 & -1 &  0 & 0 \\ 
        0 &  0 & -1 & 0 \\ 
        0 &  0 &  0 & 1 
    \end{bmatrix}
    \label{eq:coord_transform}
\end{equation}
This rigorous alignment allows analysts to maintain a consistent mental map of the 3D space, ensuring that interactive manipulations directly map to geometrically correct synthesized views.

Beyond extrinsic alignment, accurate view synthesis mandates the precise extraction of intrinsic camera parameters from the interactive viewport. Traditional explicit rendering pipelines often embed camera state management inextricably within the rasterization loop, making customized trajectory design cumbersome. 
VisEngine bypasses this by introducing a dynamic intrinsic mapping module. For each frame, the engine continuously retrieves the active viewport resolution $(W, H)$ and the horizontal Field-of-View ($\text{FoV}_x$). The intrinsic matrix $\mathbf{K}$ is then formulated by computing the focal lengths $f_x, f_y$ and projecting the principal point $(c_x, c_y)$ to the image center:
\begin{equation}
    \mathbf{K} = 
    \begin{bmatrix} 
        \frac{W}{2 \tan(\text{FoV}_x / 2)} & 0 & \frac{W}{2} \\ 
        0 & \frac{H}{2 \tan(\text{FoV}_y / 2)} & \frac{H}{2} \\ 
        0 & 0 & 1 
    \end{bmatrix}
\end{equation}
This rigorous mathematical mapping accommodates dynamic zooming and aspect-ratio modifications on the fly, which are essential operations during fine-grained spatial inspection.

\begin{figure*}[h]
  \centerline{\includegraphics[width=1.0\textwidth, trim=1.5cm 0.5cm 3.5cm 2cm, clip]{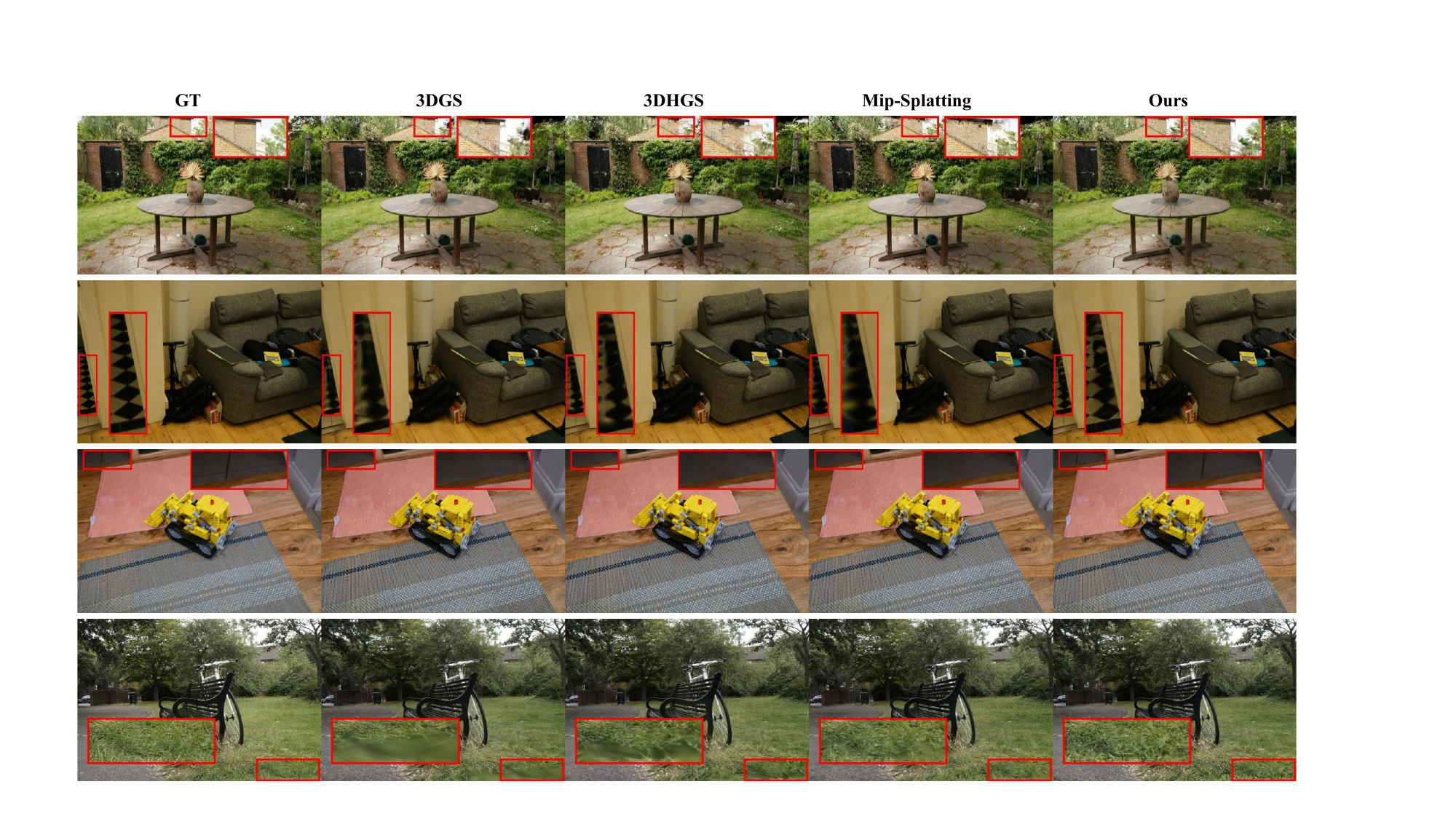}}
  \caption{\textbf{Comparison Results.} Visual differences are highlighted with red insets for better clarity. Our approach consistently outperforms other models on Mip-NeRF 360~\cite{barron2022mip} dataset, demonstrating clear advantages in challenging scenarios such as thin geometries and fine-scale details. Best viewed in color.}
\label{e1}
\end{figure*}

\begin{figure*}[h]
  \centerline{\includegraphics[width=1.0\textwidth]{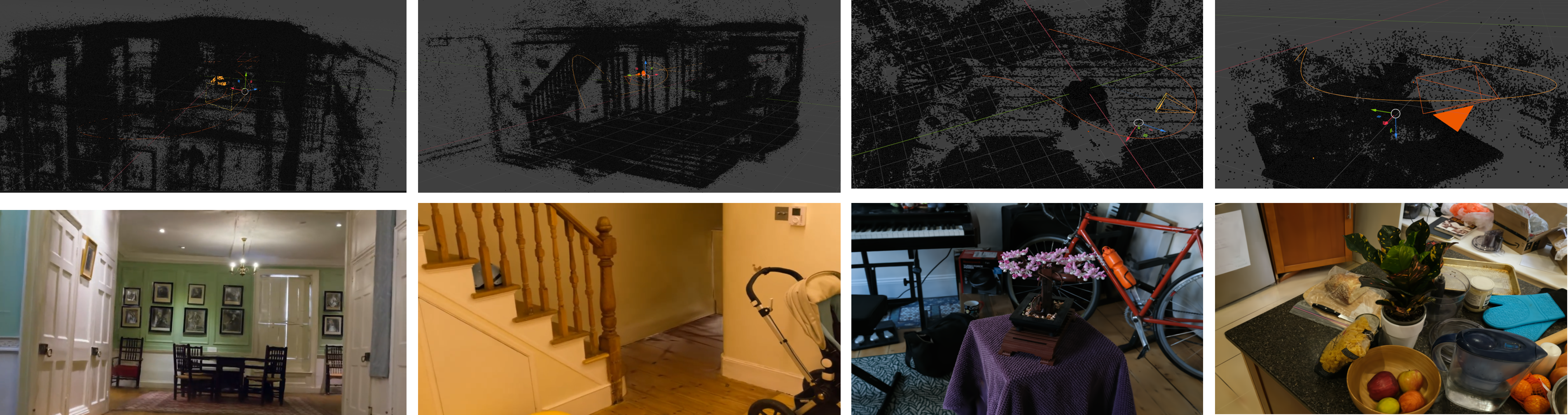}}
\caption{\textbf{Interactive Exploration and View Synthesis Quality.} 
\textit{Top row:} The unconstrained free camera trajectory and the 3DSGS point cloud proxy visualized within our Blender frontend. 
\textit{Bottom row:} The corresponding novel views rendered by our CUDA rasterizer at the selected poses. }
\label{b1}
\end{figure*}

To fundamentally realize the architectural decoupling, we designed a lightweight parameter serialization protocol. Rather than embedding the computationally heavy CUDA rasterizer directly within the frontend Python API, which would inevitably block the main UI thread and degrade the interaction frame rate.
VisEngine extracts the synchronized spatio-temporal states ($\mathbf{M}_{\text{cv}}$, $\mathbf{K}$, $W$, $H$). 
For pre-designed analytical trajectories, this sequence of matrices is serialized into a standardized, lightweight format . 
Our generalized CUDA backend asynchronously ingests this data stream to execute the forward rasterization pass.

This decoupled architecture offers profound analytical advantages. 
First, users can effortlessly navigate intricate topologies and design precise temporal visual narratives without being hindered by programming interfaces. 
Moreover, by isolating the frontend interaction loop from the backend rendering latencies, VisEngine ensures that the visual interface remains highly responsive and robust, even when evaluating millions of complex asymmetric primitives. 

\section{Experiment}

\begin{table*}[h]
\centering
\caption{\textbf{Quantitative comparison to previous methods on Mip-NeRF 360~\cite{barron2022mip}, Tanks $\&$ Temples~\cite{Knapitsch2017}, and Deep Blending~\cite{hedman2018deep} dataset.} The higher PSNR(↑) and SSIM(↑) denote better rendering quality. The color of each cell shows the \colorbox[HTML]{F09BA0}{best} and the \colorbox[HTML]{FCCF93}{second best}.}
\begin{tabular}{l|ccc|ccc|ccc}
\toprule
Dataset     & \multicolumn{3}{c|}{Mip-NeRF 360~\cite{barron2022mip}} & \multicolumn{3}{c|}{Tanks $\&$ Temples~\cite{Knapitsch2017}} & \multicolumn{3}{c}{Deep Blending~\cite{hedman2018deep}} \\
Method      & PSNR↑    & SSIM↑    & LPIPS↓  & PSNR↑    & SSIM↑    & LPIPS↓  & PSNR↑    & SSIM↑    & LPIPS↓     \\
\midrule
Mip-NeRF~\cite{barron2021mip}    & 29.23   & 0.844   & 0.207  & 22.22  & 0.759  & 0.257  & 29.40     & 0.901     & 0.245     \\
3DGS~\cite{kerbl20233d}        & 28.88   & 0.870   & 0.182  & 23.60  & 0.847  & 0.181  & 29.41     & 0.903     & 0.243     \\
GES~\cite{hamdi2024ges}         & 28.69   & 0.857   & 0.206  & 23.35  & 0.836  & 0.198  & 29.68     & 0.901     & 0.252     \\
2DGS~\cite{Huang2DGS2024}        & 28.49 & 0.862 & 0.201 & 23.11 & 0.829 & 0.212 & 29.52 & 0.899 & 0.259 \\
Fre-GS~\cite{zhang2024fregs}      & 27.85   & 0.826   & 0.209  & 23.96  & 0.841  & 0.183  & 29.73     & 0.903     & 0.245     \\
Scaffold-GS~\cite{lu2024scaffold} & 28.84   & 0.848   & 0.220  & 23.96  & 0.853  & 0.177  & \cellcolor[HTML]{FCCF93}30.21     & \cellcolor[HTML]{F09BA0}0.906     & 0.254     \\
Mip-Splatting~\cite{Yu2024MipSplatting}& 29.31  & \cellcolor[HTML]{FCCF93}{0.880}   & \cellcolor[HTML]{FCCF93}{0.168}  & 23.83  & 0.852  & 0.175  & 29.35     & 0.903     & 0.244     \\
3DHGS~\cite{li20243d}       & \cellcolor[HTML]{FCCF93}{29.56}   & {0.873}   & {0.178}  
            & \cellcolor[HTML]{FCCF93}24.35  & \cellcolor[HTML]{F09BA0}0.857  & \cellcolor[HTML]{F09BA0}0.169  & 29.64     & 0.902     & \cellcolor[HTML]{FCCF93}0.242     \\
\midrule
3DSGS~(Ours)        & \cellcolor[HTML]{F09BA0}{29.78} & \cellcolor[HTML]{F09BA0}0.888 & \cellcolor[HTML]{F09BA0}0.145 & \cellcolor[HTML]{F09BA0}24.36 & \cellcolor[HTML]{F09BA0}0.857 & 0.173 & 29.75 & \cellcolor[HTML]{FCCF93}0.904 & \cellcolor[HTML]{F09BA0}0.236 \\
Scaffold-SGS~(Ours)         & 29.32 & 0.870 & 0.182 & 24.31 & \cellcolor[HTML]{FCCF93}0.855 & \cellcolor[HTML]{FCCF93}0.171 & \cellcolor[HTML]{F09BA0}30.26 & \cellcolor[HTML]{F09BA0}0.906 & 0.244 \\
\bottomrule
\end{tabular}
\label{t1}
\end{table*}

\begin{figure*}[!ht]
  \centerline{\includegraphics[width=1.0\textwidth]{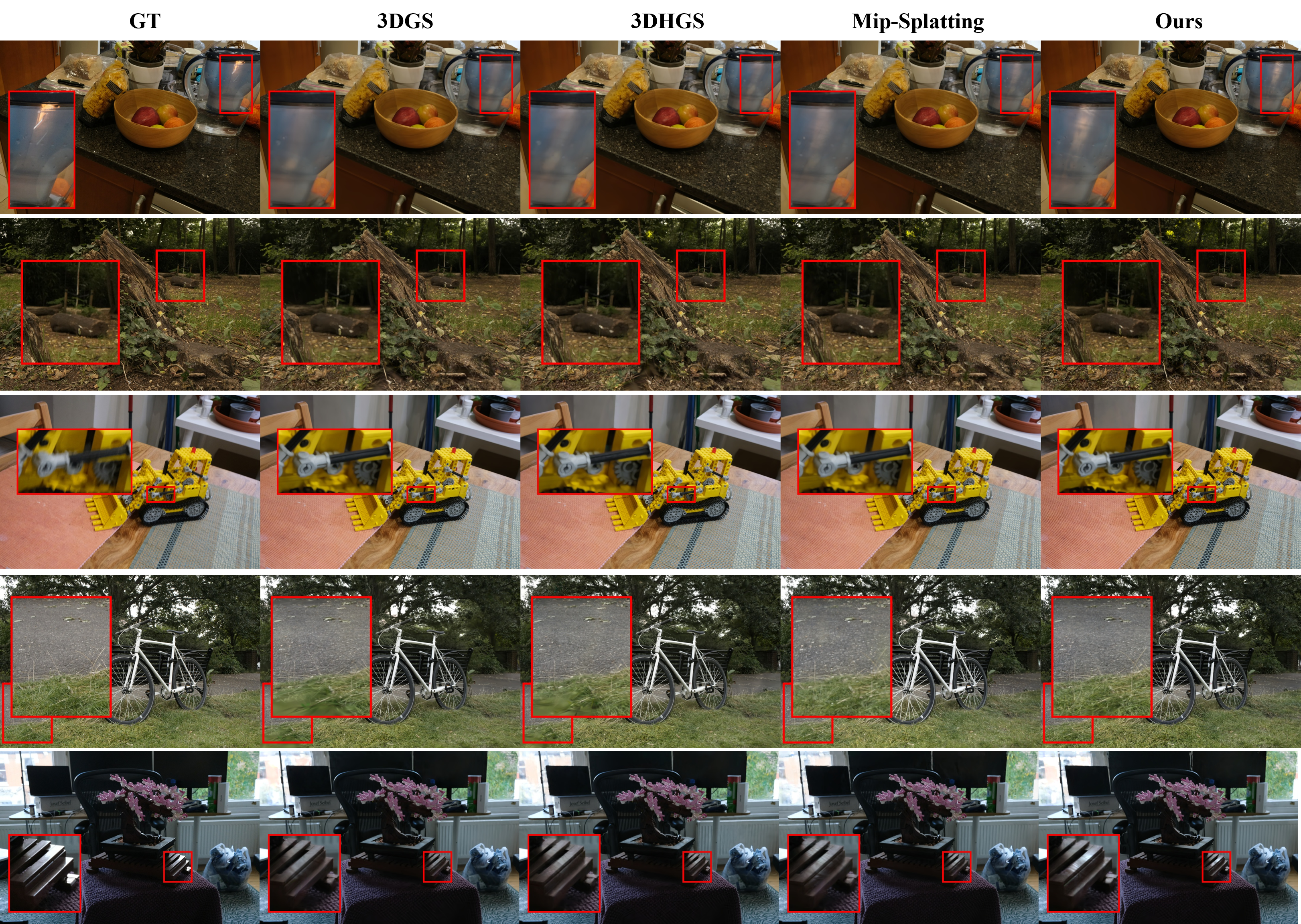}}
  \caption{\textbf{Comparison Results.} Visual differences are highlighted with red insets for better clarity. Our approach consistently outperforms other models on Mip-NeRF 360~\cite{barron2022mip} dataset, demonstrating advantages in challenging scenarios. Best viewed in color.}
\label{e3}
\end{figure*}

\begin{table*}[h]
\centering
\caption{\textbf{Training time and rendering efficiency across different datasets.}}
\begin{tabular}{lccc|ccc|ccc}
\toprule
          & \multicolumn{3}{c|}{Mip-NeRF 360~\cite{barron2022mip}}         & \multicolumn{3}{c|}{Tanks $\&$ Temples~\cite{Knapitsch2017}}           & \multicolumn{3}{c}{Deep Blending~\cite{hedman2018deep}}  \\
          \midrule
Method    & FPS & Training time & \# of kernel & FPS & Training time & \# of kernel & FPS & Training time & \# of kernel \\
3DGS~\cite{kerbl20233d}     & 99  & 21min         & 3.22M        & 120 & 14min         & 3.13M        & 104 & 21min         & 2.72M        \\
GS-MCMC~\cite{kheradmand20243d} & 75  & 32min         & 3.29M        & 98 & 20min         & 3.22M        & 90  & 29min         & 2.69M        \\
\midrule
Ours      & 91  & 49min         & 3.12M        & 112  & 31min         & 3.07M        & 101  & 38min         & 2.42M      \\
\bottomrule
\end{tabular}
\label{f1}
\end{table*}

\subsection{Experiment Setup}

\textbf{Datasets and Evaluation Metrics.} 

To rigorously evaluate the visual fidelity, structural compactness, and rendering robustness of our 3DSGS framework under unconstrained camera exploration, we conduct extensive experiments across a diverse collection of real-world scenes. In accordance with established novel view synthesis benchmarks, we utilize a comprehensive suite of 11 complex scenes sourced from three widely adopted datasets:

\begin{itemize}
\item\textbf{Mip-NeRF 360 Dataset~\cite{barron2022mip}:} We select seven high-resolution scenes (a mix of indoor and unbounded outdoor environments) from this dataset. These scenes feature complex central objects surrounded by highly detailed, expansive backgrounds.

\item\textbf{Tanks and Temples~\cite{Knapitsch2017}:} We evaluate two large-scale outdoor scenes (\textit{Truck} and \textit{Train}). This dataset is characterized by challenging, unconstrained capture trajectories, variable lighting conditions, and thin geometric structures. It serves as a rigorous stress test for our depth-aware densification strategy and the structural alignment capabilities of the skew primitives.

\item\textbf{Deep Blending~\cite{hedman2018deep}:} We incorporate two indoor scenes (\textit{Playroom} and \textit{Dr Johnson}) that contain rich textures, specular reflections, and complex object boundaries. 
\end{itemize}

\noindent\textbf{Data Splitting Protocol:} 
To ensure a fair and consistent evaluation, we strictly adhere to the standard protocol established in prior volumetric rendering research. For every scene, we uniformly sample every 8-th image from the captured sequence to form the unseen testing set, while the remaining images are utilized for training and optimizing the 3DSGS representation. 

\noindent\textbf{Quantitative Evaluation Metrics:}
To assess the quality of the synthesized novel views from both signal-processing and human-perceptual perspectives, we employ three standard quantitative metrics:

\begin{itemize}
    \item \textbf{Peak Signal-to-Noise Ratio (PSNR):} This metric measures the absolute pixel-level photometric error between the rendered image and the ground truth. A higher PSNR indicates lower pixel-wise distortion and better fundamental signal fidelity.
    \item \textbf{Structural Similarity Index Measure (SSIM)~\cite{wang2004image}:} Unlike PSNR, SSIM evaluates image quality based on the degradation of structural information, luminance, and contrast over local image patches. A higher SSIM score demonstrates that our method better preserves the geometric integrity and sharp boundaries of the objects, which is crucial for accurate spatial analysis.
    \item \textbf{Learned Perceptual Image Patch Similarity (LPIPS)~\cite{zhang2018unreasonable}:} LPIPS uses a VGG network backbone. This deep-learning-based metric measures the perceptual distance between images. A lower LPIPS indicates that rendered views contain fewer visual artifacts.
\end{itemize}

\begin{figure*}[!htp]
  \centerline{\includegraphics[width=1.0\textwidth, trim=0.5cm 14cm 7cm 0cm, clip]{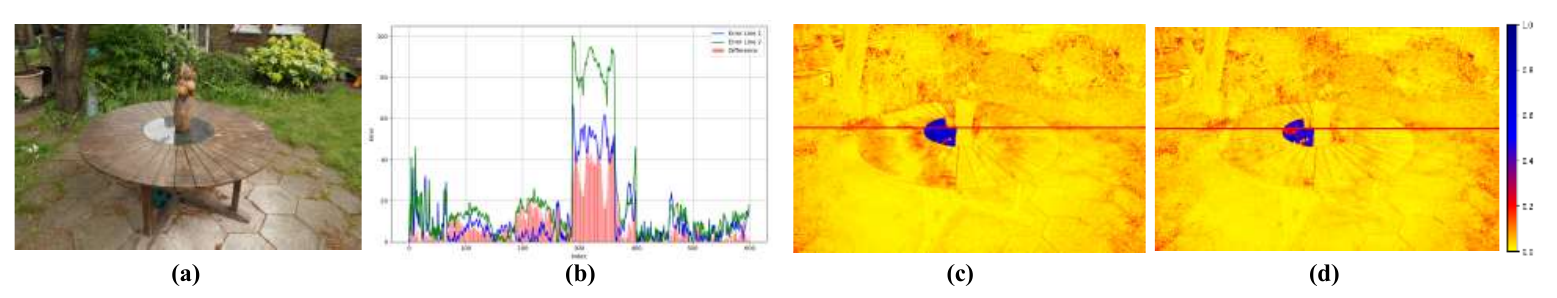}}
  \caption{This figure illustrates the fitting error analysis between different rendering models and the ground truth. 
  (a) represents the ground truth, while subfigures (c) and (d) display the error maps of our model and the 3DGS model, respectively. 
  The errors are calculated as absolute values and normalized, presented as heatmaps.
  (b) shows the error values along sampled lines on the error maps, where green indicates the error values of the 3DGS model, blue represents the error values of our model, and red highlights the differences in errors between the two models.}
\label{s3}
\end{figure*}

Detailed quantitative comparisons are provided in the Supplementary Materials.

\textbf{Baselines.} 
Given the extensive volume of research related to 3DGS, we select the original 3DGS~\cite{kerbl20233d} alongside the most recent advancements that target core improvements in the 3DGS paradigm and have demonstrated superior performance. 
The baselines include Generalized Exponential Splatting (GES)~\cite{hamdi2024ges}, Mip-Splatting~\cite{Yu2024MipSplatting} and 3D Half-Gaussian Splatting (3DHGS)~\cite{li20243d}, which leverage different (positive-only) mixture components; 
Scaffold-GS~\cite{lu2024scaffold} and Fre-GS~\cite{zhang2024fregs}, which enhance training efficiency through optimized procedures; 
2D Gaussian Splatting (2DGS) \cite{Huang2DGS2024}, which design the 2D Gaussian kernel;
and Mip-NeRF~\cite{barron2021mip}, a state-of-the-art method based on Neural Radiance Fields (NeRF)~\cite{mildenhall2021nerf}. 
Collectively, these baselines encompass techniques employing novel mixture components, optimization strategies, and represent the current state-of-the-art in rendering quality.

\subsection{Results Analysis}

\textbf{Rendering Quality Comparison:}
Quantitative evaluations are presented in Figs. \ref{e1}, along with detailed metrics summarized in Tab. \ref{t1}. 
The training time and render efficiency are evaluated in Tab. \ref{f1}.
More results could be found in the Appendix.
Across diverse datasets and rendering methodologies, the integration of the 3D Skew Gaussian Splatting (3DSGS) kernel consistently yields significant improvements.

Fig. \ref{e1} illustrates the quality of the proposed method on a complex scene from the Mip-NeRF 360~\cite{barron2022mip} dataset. 
Visual comparisons demonstrate that 3DSGS achieves a closer approximation to ground truth, especially in regions with sharp transitions and high-frequency details. Compared to baseline methods, our approach provides a more faithful representation of intricate scene features, confirming its effectiveness in capturing high-frequency content and discontinuities.

Tab. \ref{t1} showcases the performance of our model on the Tanks $\&$ Temples~\cite{Knapitsch2017} and Deep Blending~\cite{hedman2018deep} datasets. These datasets are particularly challenging due to their highly detailed textures and complex object interactions. 
Our method demonstrates superior capability in accurately rendering fine details and maintaining coherence across varying viewpoints. 
The visual results highlight the robustness of 3DSGS in handling intricate scene characteristics and discontinuities, further validating its effectiveness in diverse rendering scenarios.

Tab. \ref{t1} highlights the performance of various variants incorporating 3DSGS across multiple widely used datasets. 
The results reveal that our method surpasses existing approaches by a clear margin. 
Notably, on the Mip-NeRF 360~\cite{barron2022mip} dataset, our approach markedly improves reconstruction fidelity, leading to consistent high-quality rendering across all tested scenarios. 
Similar trends are observed on the Tanks $\&$ Temples~\cite{Knapitsch2017} and Deep Blending~\cite{hedman2018deep} datasets, where the proposed kernel yields leading performance metrics, setting new standards for neural scene representation.

\begin{figure*}[ht]
  \centerline{\includegraphics[width=1.0\textwidth]{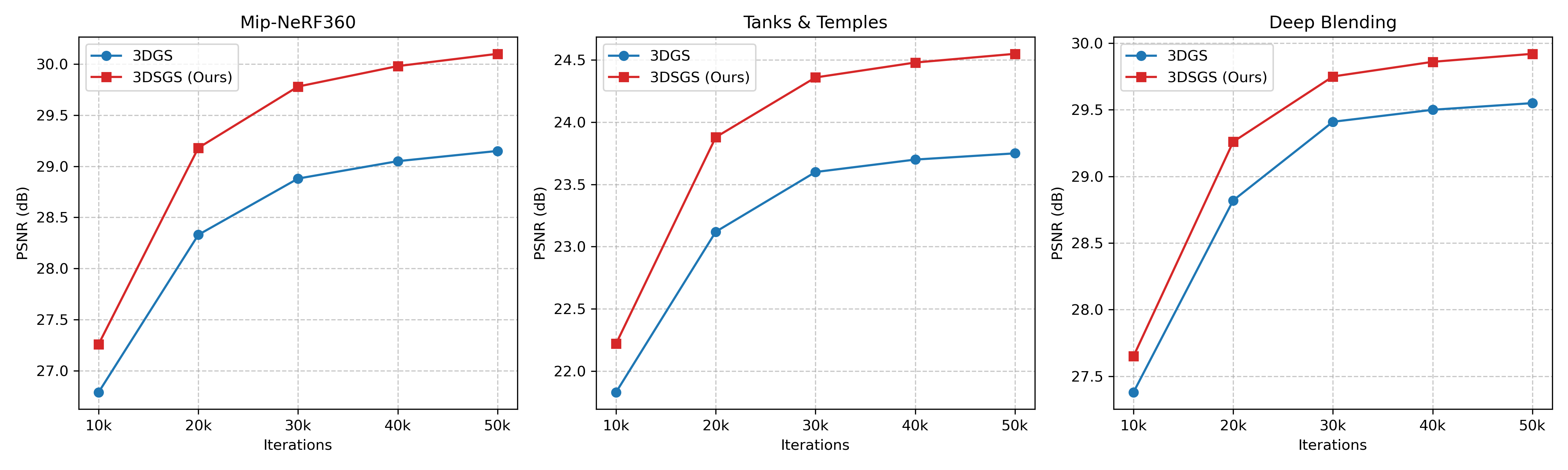}}
  \caption{\textbf{Ablation Results.} We design the ablation study about the performance of our methods with 3DGS under different optimization iteration settings across all three datasets. Our method can outperform 3DGS consistently. }
\label{p1}
\end{figure*}

The efficiency of 3DSGS is also evaluated as shown in Tab. \ref{f1}. 
The 3DSGS can achieve fewer number of kernels for 3D representation in most scenes.
Despite 3DSGS is slower compared to vanilla 3DGS, the system still maintains real-time performance at over 90 FPS.
This demonstrates that the enhanced expressiveness and detail preservation afforded by our method are achievable without compromising computational efficiency too much.

In the Fig. \ref{s3}, the fitting error analysis compares the performance of our rendering model against the 3DGS with the ground truth from the garden scene of the Mip-NeRF 360 dataset. 
Fig. \ref{s3}(a) depicts the ground truth, serving as a reference for evaluating the models’ accuracy. 
Fig. \ref{s3}(c) and (d) show the error maps of our model and the 3DGS respectively, with errors visualized as normalized heatmaps.
It is evident that our model consistently exhibits lower error magnitudes across the sampled regions, indicating more accurate reconstructions. 
Fig. \ref{s3}(b) further illustrates the error distributions along selected lines, where the blue curves (our model) generally stay below the green curves (3DGS), demonstrating reduced error levels. 
The red highlights reveal that our model minimizes discrepancies in error profiles, underscoring its robustness. 
These results highlight that our approach achieves more precise renderings and improved consistency with the ground truth compared to the 3DGS.

\noindent\textbf{Interactive Exploration and Structural Fidelity:}
Beyond static quantitative metrics, the true value of our framework is demonstrated through its robust performance during unconstrained Exploratory Data Analysis (EDA). 
As illustrated in Figure~\ref{b1}, we evaluate the system's qualitative behavior under dynamic, human-driven camera trajectories. During the frontend navigation in Blender, analysts frequently position the camera at extreme grazing angles or close to complex topological structures to inspect specific spatial phenomena. 

Standard 3DGS models exhibit visual artifacts when navigating around thin geometries. 
In contrast, our 3DSGS framework preserves these fine-scale details. 
The intrinsic skewness of our primitives naturally conforms to shape discontinuities, while the depth-aware densification ensures that primitives are compactly aligned along true physical surfaces. Crucially, the decoupled architecture of VisEngine ensures that the heavy computational evaluation of these asymmetric kernels does not block the interaction loop. Consequently, analysts can smoothly scrub through the timeline, design complex fly-throughs, and receive high-fidelity, real-time visual feedback, bridging the gap between rigorous spatial reconstruction and fluid visual analysis.

\begin{table}[t]
\centering
\caption{Ablation evaluation on the Mip-NeRF 360~\cite{barron2022mip} dataset. }
\begin{tabular}{lccc}
\toprule
  Method   & PSNR↑  & SSIM↑  & LPIPS↓ \\
     \midrule
w/o opacity regularization  & 29.53 & 0.881 & 0.159 \\
w/o densification optimization & 29.47 & 0.878 & 0.166 \\
\midrule
Ours & \textbf{29.78} & \textbf{0.888} & \textbf{0.145} \\
\bottomrule
\end{tabular}
\label{a1}
\end{table}

\subsection{Ablation Analysis}

To demonstrate the necessity of the proposed modules within our 3D Skew Gaussian Splatting framework, we conduct ablation studies focusing on the Depth-Aware Densification Criterion and opacity regularization as shown in Fig.\ref{a1}. 
The Depth-Aware Densification Criterion plays a crucial role in ensuring accurate and coherent spatial representation by adaptively concentrating the density of splats in regions with complex geometry, thereby improving the fidelity of the reconstructed scene. 
Without this module, the representation tends to be sparse in areas of rich details, leading to poor rendering quality and loss of structural consistency. 

Meanwhile, opacity regularization serves as an essential mechanism to balance the transparency and solidity of splats, encouraging the model to produce more realistic and stable visualizations. 
Removing this regularization results in overly transparent or artificially opaque regions, adversely affecting the visual coherence and physical plausibility of the output.

In addition, we present a quantitative ablation study investigating the model's performance under varying optimization iteration settings. 
As shown in Fig.~\ref{p1}, this analysis not only demonstrates the stable convergence behavior of our approach but also highlights its consistent superiority over the baseline at different training stages.

\subsection{Limitation} 
One limitation of our current approach is that the introduction of the skew parameter has led to increased training time and decreased rendering speed as shown in Tab. \ref{f1}. 
We believe this is primarily due to the higher computational cost of the erf function compared to the exponential function, which affects overall efficiency. 
Despite this, the system still maintains real-time performance at over 60 FPS. 
Moving forward, we plan to explore optimization strategies to reduce this computational overhead, further improving both training efficiency and rendering speed.

\section{Conclusion}  
\label{sec:conclusion}  

In this work, we presented 3D Skew Gaussian Splatting (3DSGS), a novel framework that fundamentally extends the core primitive of explicit radiance fields from a standard symmetric Gaussian to a generalized 3D Skew Normal distribution. 
Our approach resolves the structural artifacts and visual perception barriers inherent in standard 3DGS by: (1) introducing an intrinsic skewness parameter to accurately model complex spatial discontinuities and asymmetric features, (2) developing a spatially varying opacity representation for improved transparency modeling, and (3) refining the optimization process through a depth-aware densification criterion that guarantees geometric compactness. 

We re-derived the underlying CUDA rasterization pipeline and develop a dedicated visualization engine (VisEngine) within Blender. 
Empirical results demonstrate that our framework not only achieves superior visual fidelity but also empowers analysts with real-time, unconstrained exploratory data analysis. 
Future work may include speeding up the 3D skew splatting and exploring the extension of 3DSGS to dynamic scenes and deformable objects.

\section{Acknowledgment}
This work is supported by the National Natural Science Foundation of China (No. 62406267), Guangdong Provincial Project (No. 2024QN11X072), Guangzhou-HKUST(GZ) Joint Funding Program (No. 2025A03J3956) and Guangzhou Municipal Education Project (No. 2024312122).

\bibliographystyle{IEEEtran}
\bibliography{template}

@String(CVPR= {IEEE Conf. Comput. Vis. Pattern Recog.})

@String(TOG= {ACM Trans. Graph.})

@String(CVPR  = {CVPR})

@String(TOG   = {ACM TOG})

@article{kerbl20233d,
  title={3d gaussian splatting for real-time radiance field rendering.},
  author={Kerbl, Bernhard and Kopanas, Georgios and Leimk{\"u}hler, Thomas and Drettakis, George},
  journal={ACM Trans. Graph.},
  volume={42},
  number={4},
  pages={139--1},
  year={2023}
}

@article{zhu20253d,
  title={3D Student Splatting and Scooping},
  author={Zhu, Jialin and Yue, Jiangbei and He, Feixiang and Wang, He},
  journal={arXiv preprint arXiv:2503.10148},
  year={2025}
}

@article{li20243d,
  title={3d-hgs: 3d half-gaussian splatting},
  author={Li, Haolin and Liu, Jinyang and Sznaier, Mario and Camps, Octavia},
  journal={arXiv preprint arXiv:2406.02720},
  year={2024}
}

@article{huang2024deformable,
  title={Deformable Radial Kernel Splatting},
  author={Huang, Yi-Hua and Lin, Ming-Xian and Sun, Yang-Tian and Yang, Ziyi and Lyu, Xiaoyang and Cao, Yan-Pei and Qi, Xiaojuan},
  journal={arXiv preprint arXiv:2412.11752},
  year={2024}
}

@article{mildenhall2021nerf,
  title={Nerf: Representing scenes as neural radiance fields for view synthesis},
  author={Mildenhall, Ben and Srinivasan, Pratul P and Tancik, Matthew and Barron, Jonathan T and others},
  journal={Communications of the ACM},
  volume={65},
  number={1},
  pages={99--106},
  year={2021},
  publisher={ACM New York, NY, USA}
}

@inproceedings{barron2021mip,
  title={Mip-nerf: A multiscale representation for anti-aliasing neural radiance fields},
  author={Barron, Jonathan T and Mildenhall, Ben and Tancik, Matthew and Hedman, Peter and Martin-Brualla, Ricardo and Srinivasan, Pratul P},
  booktitle={Proceedings of the IEEE/CVF international conference on computer vision},
  pages={5855--5864},
  year={2021}
}

@article{muller2022instant,
  title={Instant neural graphics primitives with a multiresolution hash encoding},
  author={M{\"u}ller, Thomas and Evans, Alex and Schied, Christoph and Keller, Alexander},
  journal={ACM transactions on graphics (TOG)},
  volume={41},
  number={4},
  pages={1--15},
  year={2022},
  publisher={ACM New York, NY, USA}
}

@article{held20243d,
  title={3D Convex Splatting: Radiance Field Rendering with 3D Smooth Convexes},
  author={Held, Jan and Vandeghen, Renaud and Hamdi, Abdullah and Deliege, Adrien and Cioppa, Anthony and Giancola, Silvio and Vedaldi, Andrea and Ghanem, Bernard and Van Droogenbroeck, Marc},
  journal={arXiv preprint arXiv:2411.14974},
  year={2024}
}

@inproceedings{barron2022mip,
  title={Mip-nerf 360: Unbounded anti-aliased neural radiance fields},
  author={Barron, Jonathan T and Mildenhall, Ben and Verbin, Dor and Srinivasan, Pratul P and Hedman, Peter},
  booktitle={Proceedings of the IEEE/CVF conference on computer vision and pattern recognition},
  pages={5470--5479},
  year={2022}
}

@article{Knapitsch2017,
    author    = {Arno Knapitsch and Jaesik Park and Qian-Yi Zhou and Vladlen Koltun},
    title     = {Tanks and Temples: Benchmarking Large-Scale Scene Reconstruction},
    journal   = {ACM Transactions on Graphics},
    volume    = {36},
    number    = {4},
    year      = {2017},
}

@article{hedman2018deep,
  title={Deep blending for free-viewpoint image-based rendering},
  author={Hedman, Peter and Philip, Julien and Price, True and Frahm, Jan-Michael and Drettakis, George and Brostow, Gabriel},
  journal={ACM Transactions on Graphics (ToG)},
  volume={37},
  number={6},
  pages={1--15},
  year={2018},
  publisher={ACM New York, NY, USA}
}

@article{wang2004image,
  title={Image quality assessment: from error visibility to structural similarity},
  author={Wang, Zhou and Bovik, Alan C and Sheikh, Hamid R and Simoncelli, Eero P},
  journal={IEEE transactions on image processing},
  volume={13},
  number={4},
  pages={600--612},
  year={2004},
  publisher={IEEE}
}

@inproceedings{zhang2018unreasonable,
  title={The unreasonable effectiveness of deep features as a perceptual metric},
  author={Zhang, Richard and Isola, Phillip and Efros, Alexei A and Shechtman, Eli and Wang, Oliver},
  booktitle={Proceedings of the IEEE conference on computer vision and pattern recognition},
  pages={586--595},
  year={2018}
}

@inproceedings{hamdi2024ges,
  title={Ges: Generalized exponential splatting for efficient radiance field rendering},
  author={Hamdi, Abdullah and Melas-Kyriazi, Luke and Mai, Jinjie and Qian, Guocheng and Liu, Ruoshi and Vondrick, Carl and Ghanem, Bernard and Vedaldi, Andrea},
  booktitle={Proceedings of the IEEE/CVF Conference on Computer Vision and Pattern Recognition},
  pages={19812--19822},
  year={2024}
}

@inproceedings{lu2024scaffold,
  title={Scaffold-gs: Structured 3d gaussians for view-adaptive rendering},
  author={Lu, Tao and Yu, Mulin and Xu, Linning and Xiangli, Yuanbo and Wang, Limin and Lin, Dahua and Dai, Bo},
  booktitle={Proceedings of the IEEE/CVF Conference on Computer Vision and Pattern Recognition},
  pages={20654--20664},
  year={2024}
}

@inproceedings{zhang2024fregs,
  title={Fregs: 3d gaussian splatting with progressive frequency regularization},
  author={Zhang, Jiahui and Zhan, Fangneng and Xu, Muyu and Lu, Shijian and Xing, Eric},
  booktitle={Proceedings of the IEEE/CVF Conference on Computer Vision and Pattern Recognition},
  pages={21424--21433},
  year={2024}
}

@inproceedings{jiang2024gaussianshader,
  title={Gaussianshader: 3d gaussian splatting with shading functions for reflective surfaces},
  author={Jiang, Yingwenqi and Tu, Jiadong and Liu, Yuan and Gao, Xifeng and Long, Xiaoxiao and Wang, Wenping and Ma, Yuexin},
  booktitle={Proceedings of the IEEE/CVF Conference on Computer Vision and Pattern Recognition},
  pages={5322--5332},
  year={2024}
}

@inproceedings{ye20243d,
  title={3d gaussian splatting with deferred reflection},
  author={Ye, Keyang and Hou, Qiming and Zhou, Kun},
  booktitle={ACM SIGGRAPH 2024 Conference Papers},
  pages={1--10},
  year={2024}
}

@article{yang2024spec,
  title={Spec-gaussian: Anisotropic view-dependent appearance for 3d gaussian splatting},
  author={Yang, Ziyi and Gao, Xinyu and Sun, Yang-Tian and Huang, Yihua and Lyu, Xiaoyang and Zhou, Wen and Jiao, Shaohui and Qi, Xiaojuan and Jin, Xiaogang},
  journal={Advances in Neural Information Processing Systems},
  volume={37},
  pages={61192--61216},
  year={2024}
}

@inproceedings{zakharov2024human,
  title={Human hair reconstruction with strand-aligned 3d gaussians},
  author={Zakharov, Egor and Sklyarova, Vanessa and Black, Michael and Nam, Giljoo and Thies, Justus and Hilliges, Otmar},
  booktitle={European Conference on Computer Vision},
  pages={409--425},
  year={2024},
  organization={Springer}
}

@article{chen2024beyondgaussian,
  title={Beyond Gaussians: Fast and High-Fidelity 3D Splatting with Linear Kernels},
  author={Chen, Haodong and Chen, Runnan and Qu, Qiang and Wang, Zhaoqing and Liu, Tongliang and Chen, Xiaoming and Chung, Yuk Ying},
  journal={arXiv preprint arXiv:2411.12440},
  year={2024}
}

@article{xu2024supergaussians,
  title={SuperGaussians: Enhancing Gaussian Splatting Using Primitives with Spatially Varying Colors},
  author={Xu, Rui and Chen, Wenyue and Wang, Jiepeng and Liu, Yuan and Wang, Peng and Gao, Lin and Xin, Shiqing and Komura, Taku and Li, Xin and Wang, Wenping},
  journal={arXiv preprint arXiv:2411.18966},
  year={2024}
}

@article{hyung2024effective,
  title={Effective rank analysis and regularization for enhanced 3d gaussian splatting},
  author={Hyung, Junha and Hong, Susung and Hwang, Sungwon and Lee, Jaeseong and Choo, Jaegul and Kim, Jin-Hwa},
  journal={arXiv preprint arXiv:2406.11672},
  year={2024}
}

@inproceedings{rota2024revising,
  title={Revising densification in gaussian splatting},
  author={Rota Bul{\`o}, Samuel and Porzi, Lorenzo and Kontschieder, Peter},
  booktitle={European Conference on Computer Vision},
  pages={347--362},
  year={2024},
  organization={Springer}
}

@article{kheradmand20243d,
  title={3d gaussian splatting as markov chain monte carlo},
  author={Kheradmand, Shakiba and Rebain, Daniel and Sharma, Gopal and Sun, Weiwei and Tseng, Yang-Che and Isack, Hossam and Kar, Abhishek and Tagliasacchi, Andrea and Yi, Kwang Moo},
  journal={Advances in Neural Information Processing Systems},
  volume={37},
  pages={80965--80986},
  year={2024}
}

@article{kulhanek2024wildgaussians,
  title={Wildgaussians: 3d gaussian splatting in the wild},
  author={Kulhanek, Jonas and Peng, Songyou and Kukelova, Zuzana and Pollefeys, Marc and Sattler, Torsten},
  journal={arXiv preprint arXiv:2407.08447},
  year={2024}
}

@article{zha2024r,
  title={R$^2$-Gaussian: Rectifying Radiative Gaussian Splatting for Tomographic Reconstruction},
  author={Zha, Ruyi and Lin, Tao Jun and Cai, Yuanhao and Cao, Jiwen and Zhang, Yanhao and Li, Hongdong},
  journal={arXiv preprint arXiv:2405.20693},
  year={2024}
}

@InProceedings{Yu2024MipSplatting,
    author    = {Yu, Zehao and Chen, Anpei and Huang, Binbin and Sattler, Torsten and Geiger, Andreas},
    title     = {Mip-Splatting: Alias-free 3D Gaussian Splatting},
    booktitle = {Proceedings of the IEEE/CVF Conference on Computer Vision and Pattern Recognition (CVPR)},
    month     = {June},
    year      = {2024},
    pages     = {19447-19456}
}

@inproceedings{Huang2DGS2024,
    title={2D Gaussian Splatting for Geometrically Accurate Radiance Fields},
    author={Huang, Binbin and Yu, Zehao and Chen, Anpei and Geiger, Andreas and Gao, Shenghua},
    publisher = {Association for Computing Machinery},
    booktitle = {SIGGRAPH 2024 Conference Papers},
    year      = {2024},
    doi       = {10.1145/3641519.3657428}
}

@inproceedings{wen2023bundlesdf,
  title={Bundlesdf: Neural 6-dof tracking and 3d reconstruction of unknown objects},
  author={Wen, Bowen and Tremblay, Jonathan and Blukis, Valts and Tyree, Stephen and M{\"u}ller, Thomas and Evans, Alex and Fox, Dieter and others},
  booktitle={Proceedings of the IEEE/CVF Conference on Computer Vision and Pattern Recognition},
  pages={606--617},
  year={2023}
}

@article{wang2025look,
  title={Look at the Sky: Sky-aware Efficient 3D Gaussian Splatting in the Wild},
  author={Wang, Yuze and Wang, Junyi and Gao, Ruicheng and Qu, Yansong and Duan, Wantong and Yang, Shuo and Qi, Yue},
  journal={IEEE Transactions on Visualization and Computer Graphics},
  year={2025},
  publisher={IEEE}
}

@article{zhai2025splatloc,
  title={Splatloc: 3d gaussian splatting-based visual localization for augmented reality},
  author={Zhai, Hongjia and Zhang, Xiyu and Zhao, Boming and Li, Hai and He, Yijia and Cui, Zhaopeng and others},
  journal={IEEE Transactions on Visualization and Computer Graphics},
  year={2025},
  publisher={IEEE}
}

@inproceedings{choi2022time,
  title={Time-multiplexed neural holography: a flexible framework for holographic near-eye displays with fast heavily-quantized spatial light modulators},
  author={Choi, Suyeon and Gopakumar, Manu and Peng, Yifan and Kim, Jonghyun and O'Toole and others},
  booktitle={ACM SIGGRAPH 2022 Conference Proceedings},
  pages={1--9},
  year={2022}
}

@inproceedings{tang2024sparseocc,
  title={Sparseocc: Rethinking sparse latent representation for vision-based semantic occupancy prediction},
  author={Tang, Pin and Wang, Zhongdao and Wang, Guoqing and Zheng, Jilai and Ren, Xiangxuan and others},
  booktitle={Proceedings of the IEEE/CVF Conference on Computer Vision and Pattern Recognition},
  pages={15035--15044},
  year={2024}
}

@inproceedings{yang2024deformable,
  title={Deformable 3d gaussians for high-fidelity monocular dynamic scene reconstruction},
  author={Yang, Ziyi and Gao, Xinyu and Zhou, Wen and Jiao, Shaohui and Zhang, Yuqing and Jin, Xiaogang},
  booktitle={Proceedings of the IEEE/CVF conference on computer vision and pattern recognition},
  pages={20331--20341},
  year={2024}
}

@article{chung2023luciddreamer,
  title={Luciddreamer: Domain-free generation of 3d gaussian splatting scenes},
  author={Chung, Jaeyoung and Lee, Suyoung and Nam, Hyeongjin and Lee, Jaerin and Lee, Kyoung Mu},
  journal={arXiv preprint arXiv:2311.13384},
  year={2023}
}

@article{xu2024wild,
  title={Wild-gs: Real-time novel view synthesis from unconstrained photo collections},
  author={Xu, Jiacong and Mei, Yiqun and Patel, Vishal},
  journal={Advances in Neural Information Processing Systems},
  volume={37},
  pages={103334--103355},
  year={2024}
}

@inproceedings{zhou2024dreamscene360,
  title={Dreamscene360: Unconstrained text-to-3d scene generation with panoramic gaussian splatting},
  author={Zhou, Shijie and Fan, Zhiwen and Xu, Dejia and Chang, Haoran and Chari, Pradyumna and Bharadwaj, Tejas and You, Suya and Wang, Zhangyang and Kadambi, Achuta},
  booktitle={European Conference on Computer Vision},
  pages={324--342},
  year={2024},
  organization={Springer}
}

@article{cao20243dgs,
  title={3dgs-det: Empower 3d gaussian splatting with boundary guidance and box-focused sampling for 3d object detection},
  author={Cao, Yang and Jv, Yuanliang and Xu, Dan},
  journal={arXiv preprint arXiv:2410.01647},
  year={2024}
}

@inproceedings{sun2024f3dgs,
  title={F-3dgs: Factorized coordinates and representations for 3d gaussian splatting},
  author={Sun, Xiangyu and Lee, Joo Chan and Rho, Daniel and Ko, Jong Hwan and Ali, Usman and Park, Eunbyung},
  booktitle={Proceedings of the 32nd ACM International Conference on Multimedia},
  pages={7957--7965},
  year={2024}
}

@inproceedings{morgenstern2024compact,
  title={Compact 3d scene representation via self-organizing gaussian grids},
  author={Morgenstern, Wieland and Barthel, Florian and Hilsmann, Anna and Eisert, Peter},
  booktitle={European Conference on Computer Vision},
  pages={18--34},
  year={2024},
  organization={Springer}
}

@article{ruckert2022adop,
  title={{ADOP}: Approximate Differentiable One-Pixel Point Rendering},
  author={R{\"u}ckert, Darius and Franke, Linus and Stamminger, Marc},
  journal={ACM Transactions on Graphics (TOG)},
  volume={41},
  number={4},
  pages={1--14},
  year={2022},
  publisher={ACM New York, NY, USA}
}

@inproceedings{yu2024mip,
  title={Mip-Splatting: Alias-free 3D Gaussian Splatting},
  author={Yu, Zehao and Chen, Anpei and Huang, Binbin and Sattler, Torsten and Geiger, Andreas},
  booktitle={Proceedings of the IEEE/CVF Conference on Computer Vision and Pattern Recognition},
  pages={19447--19456},
  year={2024}
}

@inproceedings{huang20242dgs,
  title={{2D Gaussian Splatting} for Geometrically Accurate Radiance Fields},
  author={Huang, Binbin and Yu, Zehao and Chen, Anpei and Geiger, Andreas and Gao, Shenghua},
  booktitle={ACM SIGGRAPH 2024 Conference Papers},
  pages={1--11},
  year={2024}
}
\appendix

\section{Ethics Statement}
This work adheres to the IEEE Code of Ethics. In this study, no human subjects or animal experimentation was involved. All datasets used were sourced in compliance with relevant usage guidelines, ensuring no violation of privacy. We have taken care to avoid any biases or discriminatory outcomes in our research process. No personally identifiable information was used, and no experiments were conducted that could raise privacy or security concerns. We are committed to maintaining transparency and integrity throughout the research process.

\section{Reproducibility Statement}
We have made every effort to ensure that the results presented in this paper are reproducible. All code and datasets will be made publicly available after the paper is accepted to facilitate replication and verification. The experimental setup, including training steps, model configurations, and hardware details, is described in detail in the paper. 
We believe these measures will enable other researchers to reproduce our work and further advance the field.

\section{LLM Usage Statement}
Large Language Models (LLMs) were used to for editing and grammar enhancement in the writing and polishing of the manuscript. 

It is important to note that the LLM was not involved in the ideation, research methodology, or experimental design. All research concepts, ideas, and analyses were developed and conducted by the authors. The contributions of the LLM were solely focused on improving the linguistic quality of the paper, with no involvement in the scientific content or data analysis.

We have ensured that the LLM-generated text adheres to ethical guidelines and does not contribute to plagiarism or scientific misconduct.

\section{Detailed Derivations}

In this section, we will present a comprehensive mathematical derivation of the 3D Skew Gaussian Surface Generation (3DSGS) method. 
Given the different statistical distributions involved, it is essential to implement modifications to both the forward and backward propagation processes. 
These corrections are necessary to ensure accurate training convergence and high-quality rendering results, aligning the optimized parameters with the intended skew Gaussian models.

\subsection{Forward Pass}
3DGS is based on Gaussian distribution, also known as Normal distribution, to achieve differentiable rendering,
The distribution of the original 3DGS is as follows:

\begin{equation}
    G_i(\mathbf{x}) = \exp\left(-\frac{1}{2}(\mathbf{x}-\boldsymbol{\mu}_i)^\top \boldsymbol{\Sigma}_i^{-1} (\mathbf{x}-\boldsymbol{\mu}_i)\right)
\end{equation}

As a comparison, the distribution form of 3D SGS is different, mainly due to the introduction of an additional product term $\Phi$ to achieve skewed representation:

\begin{align}
S(\mathbf{x}) 
&= 2 G(\mathbf{x}; \boldsymbol{\mu}, \boldsymbol{\Sigma}) \Phi\left( \boldsymbol{\beta}^T  (\mathbf{x} - \boldsymbol{\mu}) \right) 
\end{align}

where $\Phi$ is the cumulative distribution function (CDF) of the standard normal distribution, which represents a linear combination of the skewness vector and the deviation from the mean.
In addition, $\Phi$ can be expanded using the $\operatorname{erf}$ function to obtain the following equation:

\begin{equation}
\Phi(z) = \frac{1}{2} \left[ 1 + \operatorname{erf}\left(\frac{z}{\sqrt{2}}\right) \right]
\end{equation}

The $\operatorname{erf}$ (error function) is a special function widely used in probability, statistics, and partial differential equations. 
It is defined by the integral:

\begin{equation}
    \operatorname{erf}(x) = \frac{2}{\sqrt{\pi}} \int_0^x e^{-t^2} \, dt
\end{equation}

The Taylor expansion of this equation yields:

\begin{equation}
\operatorname{erf}(z) = \frac{2}{\sqrt{\pi}} \left( z - \frac{z^3}{3} + \frac{z^5}{10} - \frac{z^7}{42} + \cdots \right), \quad |z| < \infty
\end{equation}

Based on this, we can expand and simplify the distribution form of 3D SGS to obtain the following form:

\begin{align}
S(\mathbf{x}) 
&= 2 G(\mathbf{x}; \boldsymbol{\mu}, \boldsymbol{\Sigma}) \Phi\left( \boldsymbol{\beta}^T  (\mathbf{x} - \boldsymbol{\mu}) \right) \\
&= 2 G(\mathbf{x}; \boldsymbol{\mu}, \boldsymbol{\Sigma}) \cdot \frac{1}{2} \left[ 1 + \operatorname{erf}\left( \frac{\boldsymbol{\beta}^T (\mathbf{x} - \boldsymbol{\mu})}{\sqrt{2}} \right) \right] \\
&= G(\mathbf{x}; \boldsymbol{\mu}, \boldsymbol{\Sigma}) \cdot \left[ 1 + \operatorname{erf}\left( \frac{\boldsymbol{\beta}^T (\mathbf{x} - \boldsymbol{\mu})}{\sqrt{2}} \right) \right] 
\end{align}

where, $\beta$ is the skewness coefficient.

{
\small
\begin{align}
\alpha
&=o_i\cdot S_i(x) = o_i\cdot\sqrt{\frac{\det(\Sigma^{\prime})}{\det(\Sigma^{\prime}+sI)}} G_i(x) \cdot \Phi\left( {\beta}^T  ({x} - {\mu}) \right)\\
&=o_i\cdot\sqrt{\frac{\det(\Sigma^{\prime})}{\det(\Sigma^{\prime}+sI)}}\exp(-\frac{1}{2}(x-\mu)^T(\Sigma^{\prime}+sI)^{-1}(x-\mu)) \\
&\cdot \left[ 1 + \operatorname{erf}\left( \frac{{\beta}^T ({x} - {\mu})}{\sqrt{2}} \right) \right]
\end{align}
}

For opacity, 2D covariance matrix, mean, and skewness coefficient, we have:

\begin{equation}
x-\mu = \begin{bmatrix}\delta_x\\\delta_y\end{bmatrix}, \beta = \begin{bmatrix}\beta_x\\\beta_y\end{bmatrix},(\Sigma^{\prime}+sI)^{-1}=\begin{bmatrix}a&b\\b&c\end{bmatrix},d=o\cdot\sqrt{\frac{\det(\Sigma^{\prime})}{\det(\Sigma^{\prime}+sI)}}=o^{\prime}
\end{equation}

According to the above equation, the calculation of $\alpha$ can be simplified:

{
\small
\begin{align}
\alpha
&=(o_i\cdot\sqrt{\frac{\det(\Sigma^{\prime})}{\det(\Sigma^{\prime}+sI)}})\cdot\exp(-\frac{1}{2}\begin{bmatrix}\delta_x,\delta_y\end{bmatrix}\begin{bmatrix}a&b\\b&c\end{bmatrix}\begin{bmatrix}\delta_x\\\delta_y\end{bmatrix}) \\
&\cdot 
\left[ 1 + \operatorname{erf}\left( \frac{\begin{bmatrix}\beta_x,\beta_y\end{bmatrix}\begin{bmatrix}\delta_x\\\delta_y\end{bmatrix}}{\sqrt{2}} \right) \right]\\
&=d\cdot\exp(-\frac{1}{2}(a\delta_x^2+c\delta_y^2)-b\delta_x\delta_y) \cdot \left[ 1 + \operatorname{erf}\left( \frac{\beta_x\delta_x+\beta_x\delta_y}{\sqrt{2}} \right) \right]
\end{align}
}

Then, based on alpha-blending, we can calculate the color values rendered at each pixel:

\begin{equation}
C=\sum_{i=1}^Nc_i\alpha_iT_i, \quad where\quad T_i=\prod_{j=1}^{i-1}(1-\alpha_j),
\end{equation}

\subsection{Backward Pass}

In the process of backward, we need to calculate various key gradients, including $\frac{\partial L}{\partial\mu},\frac{\partial L}{\partial S},\frac{\partial L}{\partial R},\frac{\partial L}{\partial c},\frac{\partial L}{\partial o}$ and $\frac{\partial L}{\partial\beta}$.
According to the chain rule, we can obtain $\frac{\partial L}{\partial\mu},\frac{\partial L}{\partial S},\frac{\partial L}{\partial R},\frac{\partial L}{\partial c},\frac{\partial L}{\partial o}$ by calculating $\frac{\partial L}{\partial x_{ndc}}, \frac{\partial L}{\partial y_{ndc}}, \frac{\partial L}{\partial\alpha_k}, \frac{\partial L}{\partial a}, \frac{\partial L}{\partial b}, \frac{\partial L}{\partial c}$, so we will focus on the derivation of $\frac{\partial L}{\partial x_{ndc}}, \frac{\partial L}{\partial y_{ndc}}, \frac{\partial L}{\partial\alpha_k}, \frac{\partial L}{\partial a}, \frac{\partial L}{\partial b}, \frac{\partial L}{\partial c}$. We will not repeat the theoretical derivation of the remaining backpropagation here.

The loss $L$ is calculated between the ground-truth pixel colors and rendered colors.
It is obvious that $\partial C(p)/\partial c_k=\alpha_kT_k$, then

\begin{equation}\frac{\partial L}{\partial c_k}=\sum_p\frac{\partial L}{\partial C(p)}\frac{\partial C(p)}{\partial c_k}=\sum_p\alpha_kT_k\frac{\partial L}{\partial C(p)}\end{equation}

If $[i]$ is the channel index of the color, then

\begin{equation}
\frac{\partial L}{\partial\alpha_k}=\sum_i\frac{\partial L}{\partial C[i]}\frac{\partial C[i]}{\partial\alpha_k}=T_k\sum_i\frac{\partial L}{\partial C[i]}(c_k[i]-R_{k+1})
\end{equation}

According to the last section, we have:

\begin{align}
\alpha=o^{\prime}\cdot S^{\prime}(x)&=d\cdot\exp(-\frac{1}{2}(a\delta_x^2+c\delta_y^2)-b\delta_x\delta_y) \\
&\cdot \left[ 1 + \operatorname{erf}\left( \frac{\beta_x\delta_x+\beta_x\delta_y}{\sqrt{2}} \right) \right]
\end{align}

The derivative calculation formula of erf function is as follows:

\begin{equation}
\frac{d}{dx} \operatorname{erf}(x) = \frac{2}{\sqrt{\pi}} e^{-x^2}
\end{equation}

According to the derivative of erf function, we have:

\begin{align} 
\frac{d \operatorname{erf}(z)}{dz}  
&= \frac{2}{\sqrt{\pi}} e^{-z^2} \cdot \frac{dz}{dx} \\
&= \frac{2}{\sqrt{\pi}} e^{-z^2} \cdot \frac{1}{\sqrt{2}}  {\beta} \\
&= \sqrt{\frac{{2}}{{\pi}}} e^{-z^2}  {\beta} 
\end{align}

Due to
\begin{equation}
x_{ndc}=(x-\frac{W}{2})/\frac{W}{2}=\frac{2x}{W}-1
\end{equation}

Then

\begin{equation}
\delta_{x}=x-u=(x_{ndc}+1)\cdot\frac{W}{2}-u,
\end{equation}

So we have

\begin{equation}
\frac{\partial\delta_x}{\partial x_{ndc}}=\frac{W}{2}
\end{equation}

{
\small
\begin{align}
\frac{\partial \alpha}{\partial x_{ndc}}
&= \frac{\partial\alpha}{\partial\delta_x}\frac{\partial\delta_x}{\partial x_{ndc}} = d \cdot \frac{\partial S^{\prime}}{\partial\delta_x}\frac{\partial\delta_x}{\partial x_{ndc}}\\
&= -d \cdot \exp\left( -\frac{1}{2}(a\delta_x^2 + c\delta_y^2) - b\delta_x\delta_y \right) \cdot \left( 1 + \operatorname{erf}\left( \frac{\beta_x\delta_x + \beta_y\delta_y}{\sqrt{2}} \right) \right) \\
&\cdot (a\delta_x + b\delta_y) \cdot \frac{\partial \delta_x}{\partial x_{ndc}} \\
& + d \cdot \exp\left( -\frac{1}{2}(a\delta_x^2 + c\delta_y^2) - b\delta_x\delta_y \right) \cdot  \frac{\sqrt{2} \beta_x}{\sqrt{\pi}} \exp\left( -\frac{(\beta_x\delta_x + \beta_y\delta_y)^2}{2} \right)  \\
&= -d \cdot \exp\left( -\frac{1}{2}(a\delta_x^2 + c\delta_y^2) - b\delta_x\delta_y \right) \cdot \frac{\partial \delta_x}{\partial x_{ndc}} \cdot \Bigg[ (a\delta_x + b\delta_y) \\ 
&\cdot \left( 1 + \operatorname{erf}\left( \frac{\beta_x\delta_x + \beta_y\delta_y}{\sqrt{2}} \right) \right) \nonumber \quad - \frac{\sqrt{2} \beta_x}{\sqrt{\pi}} \exp\left( -\frac{(\beta_x\delta_x + \beta_y\delta_y)^2}{2} \right) \Bigg] \\
&= -d \cdot \exp\left( -\frac{1}{2}(a\delta_x^2 + c\delta_y^2) - b\delta_x\delta_y \right) \cdot \frac{\partial \delta_x}{\partial x_{ndc}} \cdot \Bigg[ (a\delta_x + b\delta_y) \\ 
&\cdot \left( 1 + \operatorname{erf}(z) \right) \nonumber \quad - \frac{\sqrt{2} \beta_x}{\sqrt{\pi}} \exp\left( -\frac{z^2}{2} \right) \Bigg] \\
&= -d \cdot G^{\prime}(x) \cdot \frac{\partial \delta_x}{\partial x_{ndc}} \cdot \Bigg[ (a\delta_x + b\delta_y) \cdot \left( 1 + \operatorname{erf}(z) \right) \nonumber- \frac{\sqrt{2} \beta_x}{\sqrt{\pi}} \exp\left( -\frac{z^2}{2} \right) \Bigg] \\
&= -d \cdot G^{\prime}(x) \cdot \frac{ W}{ 2} \cdot \Bigg[ (a\delta_x + b\delta_y) \cdot \left( 1 + \operatorname{erf}(z) \right) \nonumber- \frac{\sqrt{2} \beta_x}{\sqrt{\pi}} \exp\left( -\frac{z^2}{2} \right) \Bigg] \\
&= -d \cdot G^{\prime}(x) \cdot \frac{ W}{ 2} \cdot \Bigg[ (a\delta_x + b\delta_y) \cdot \left( 1 + \operatorname{erf}(z) \right) \nonumber- \sqrt{\frac{{2} }{{\pi}}} \beta_x \exp\left( -\frac{z^2}{2} \right) \Bigg] 
\end{align}
}

where $z = \beta_x\delta_x + \beta_y\delta_y$.

Therefore,

\begin{align}
\frac{\partial L}{\partial x_{ndc}}
&=\frac{\partial L}{\partial\alpha}\frac{\partial\alpha}{\partial\delta_x}\frac{\partial\delta_x}{\partial x_{ndc}}\\
&=-d \cdot \frac{\partial L}{\partial\alpha} \cdot G^{\prime}(x) \cdot \frac{ W}{ 2} \cdot \Bigg[ (a\delta_x + b\delta_y) \cdot \left( 1 + \operatorname{erf}(z) \right) \\
&\nonumber+ \sqrt{\frac{{2} }{{\pi}}} \beta_x \exp\left( -\frac{z^2}{2} \right) \Bigg]
\end{align}

Similarly,

\begin{align}
\frac{\partial L}{\partial y_{ndc}}
&=\frac{\partial L}{\partial\alpha}\frac{\partial\alpha}{\partial\delta_y}\frac{\partial\delta_y}{\partial y_{ndc}}\\
&=-d \cdot \frac{\partial L}{\partial\alpha} \cdot G^{\prime}(x) \cdot \frac{H}{ 2} \cdot \Bigg[ (b\delta_x + c\delta_y) \cdot \left( 1 + \operatorname{erf}(z) \right) \\
&\nonumber+ \sqrt{\frac{{2} }{{\pi}}} \beta_y \exp\left( -\frac{z^2}{2} \right) \Bigg]
\end{align}

For the derivation of $(\Sigma^{\prime}+sI)^{-1}$ and opacity, we have:

\begin{equation}
\frac{\partial L}{\partial a}=\frac{\partial L}{\partial \alpha}\frac{\partial \alpha}{\partial a}=-\frac{1}{2}d\cdot\delta_x^2\cdot S^{\prime}(x)\frac{\partial L}{\partial\alpha_k},
\end{equation}

\begin{equation}
\frac{\partial L}{\partial b}=\frac{\partial L}{\partial \alpha}\frac{\partial \alpha}{\partial b}=-d\cdot\delta_x\delta_y\cdot S^{\prime}(x)\frac{\partial L}{\partial\alpha_k},
\end{equation}

\begin{equation}
\frac{\partial L}{\partial c}=\frac{\partial L}{\partial \alpha}\frac{\partial \alpha}{\partial c}=-\frac{1}{2}d\cdot\delta_y^2\cdot S^{\prime}(x)\frac{\partial L}{\partial\alpha_k}
\end{equation}

\begin{equation}
\frac{\partial L}{\partial d}=\frac{\partial L}{\partial \alpha}\frac{\partial \alpha}{\partial d}=S^{\prime}(x)\frac{\partial L}{\partial\alpha_k} \cdot \beta_{reg}
\end{equation}

where $\beta_{reg}$ is the skewness regularization term for opacity.
For the derivation of skewness coefficient, we have:

\begin{equation}
\frac{\partial L}{\partial \boldsymbol{\beta}}=\frac{\partial L}{\partial \alpha_k} \cdot \frac{\partial \alpha_k}{\partial S} \cdot \frac{\partial S}{\partial \beta}
\end{equation}

According to

\begin{equation}
\frac{\partial S(\mathbf{x})}{\partial \boldsymbol{\beta}} = G(\mathbf{x}; \boldsymbol{\mu}, \boldsymbol{\Sigma}) \cdot \frac{2}{\sqrt{\pi}} \cdot e^{ - \left( \frac{\boldsymbol{\beta}^T (\mathbf{x} - \boldsymbol{\mu})}{\sqrt{2}} \right)^2 } \cdot \frac{1}{\sqrt{2}} (\mathbf{x} - \boldsymbol{\mu})
\end{equation}

Then

\begin{align}
\frac{\partial S}{\partial \boldsymbol{\beta_x}}
&= G^{\prime}(x) \frac{2}{\sqrt{\pi}} \cdot \exp\left({ - \left( \frac{\boldsymbol{\beta}_x \delta_x + \boldsymbol{\beta}_x \delta_x}{\sqrt{2}} \right)^2 }\right) \cdot \frac{1}{\sqrt{2}} \delta_x\\
&= \exp(-\frac{1}{2}(a\delta_x^2+c\delta_y^2)-b\delta_x\delta_y) \frac{2}{\sqrt{\pi}} \\
&\cdot \exp\left({ - \left( \frac{\boldsymbol{\beta}_x \delta_x + \boldsymbol{\beta}_x \delta_x}{\sqrt{2}} \right)^2 }\right) \cdot \frac{1}{\sqrt{2}} \delta_x\\
&= \sqrt{\frac{2}{\pi}}\delta_x \exp(-\frac{1}{2}(a\delta_x^2+c\delta_y^2 + \left( \boldsymbol{\beta}_x \delta_x + \boldsymbol{\beta}_x \delta_x \right)^2)-b\delta_x\delta_y)
\end{align}

Similarly, 

\begin{align}
\frac{\partial S}{\partial \boldsymbol{\beta_y}}
&= \sqrt{\frac{2}{\pi}}\delta_y \exp(-\frac{1}{2}(a\delta_x^2+c\delta_y^2 + \left( \boldsymbol{\beta}_x \delta_x + \boldsymbol{\beta}_x \delta_x \right)^2)-b\delta_x\delta_y)
\end{align}

Therefore,

\begin{align}
\frac{\partial L}{\partial \boldsymbol{\beta_x}}
&= d \cdot \frac{\partial L}{\partial \alpha_k} \sqrt{\frac{2}{\pi}}\delta_x \exp(-\frac{1}{2}(a\delta_x^2+c\delta_y^2 + \left( \boldsymbol{\beta}_x \delta_x + \boldsymbol{\beta}_x \delta_x \right)^2)-b\delta_x\delta_y)
\end{align}

\begin{align}
\frac{\partial L}{\partial \boldsymbol{\beta_y}}
&= d \cdot \frac{\partial L}{\partial \alpha_k} \sqrt{\frac{2}{\pi}}\delta_y \exp(-\frac{1}{2}(a\delta_x^2+c\delta_y^2 + \left( \boldsymbol{\beta}_x \delta_x + \boldsymbol{\beta}_x \delta_x \right)^2)-b\delta_x\delta_y)
\end{align}

\section{Detailed Experiment Results}

In this section, we analyze the rendering speed, training duration, and the number of kernels used to represent scenes in our proposed model. We provide a comprehensive comparison to highlight the efficiency and scalability of our approach. 
Additionally, we discuss the limitations of our method, including potential challenges in handling highly complex scenes and scalability issues under certain conditions. 
To support our analysis, we present detailed experimental results for each dataset and scene, including quantitative metrics and extensive visualizations that illustrate our model’s performance and behavior across different scenarios.

\subsection{Implementation} 
Regarding the experimental results, some baseline performances reported in the literature are taken directly from their respective publications. 
To facilitate comprehensive comparison, we also implement these methods under consistent experimental settings based on original 3DGS to ensure fairness and comparability. 

In our implementation, we extend the 3DGS framework by incorporating skewness parameters, which results in only a minimal increase in memory.
The modifications to both the forward and backward passes of the rasterizer are based on the standard 3DGS, ensuring seamless integration. 
For the 3DHGS, we follow the training protocols and hyperparameters.
For Scaffold-HGS, we do not increase the number of parameters within each Gaussian; instead, we employ an MLP to generate the normal vector for each voxel based on its feature vector, doubling the width of the output layer for the opacity MLP to predict two opacity values per voxel. 
The rasterizer is similarly adapted based on the specifications in the referenced works, ensuring consistency across different model variants.
The learning rate of skewness parameter is set to 0.0001. 
For the \( \tau_{uv} \), it is set to 0.001 by default.
All experiments are conducted using an NVIDIA RTX A6000 GPU.

\begin{figure*}[ht]
  \centerline{\includegraphics[width=1.0\textwidth, trim=0cm 3.5cm 4.5cm 0cm, clip]{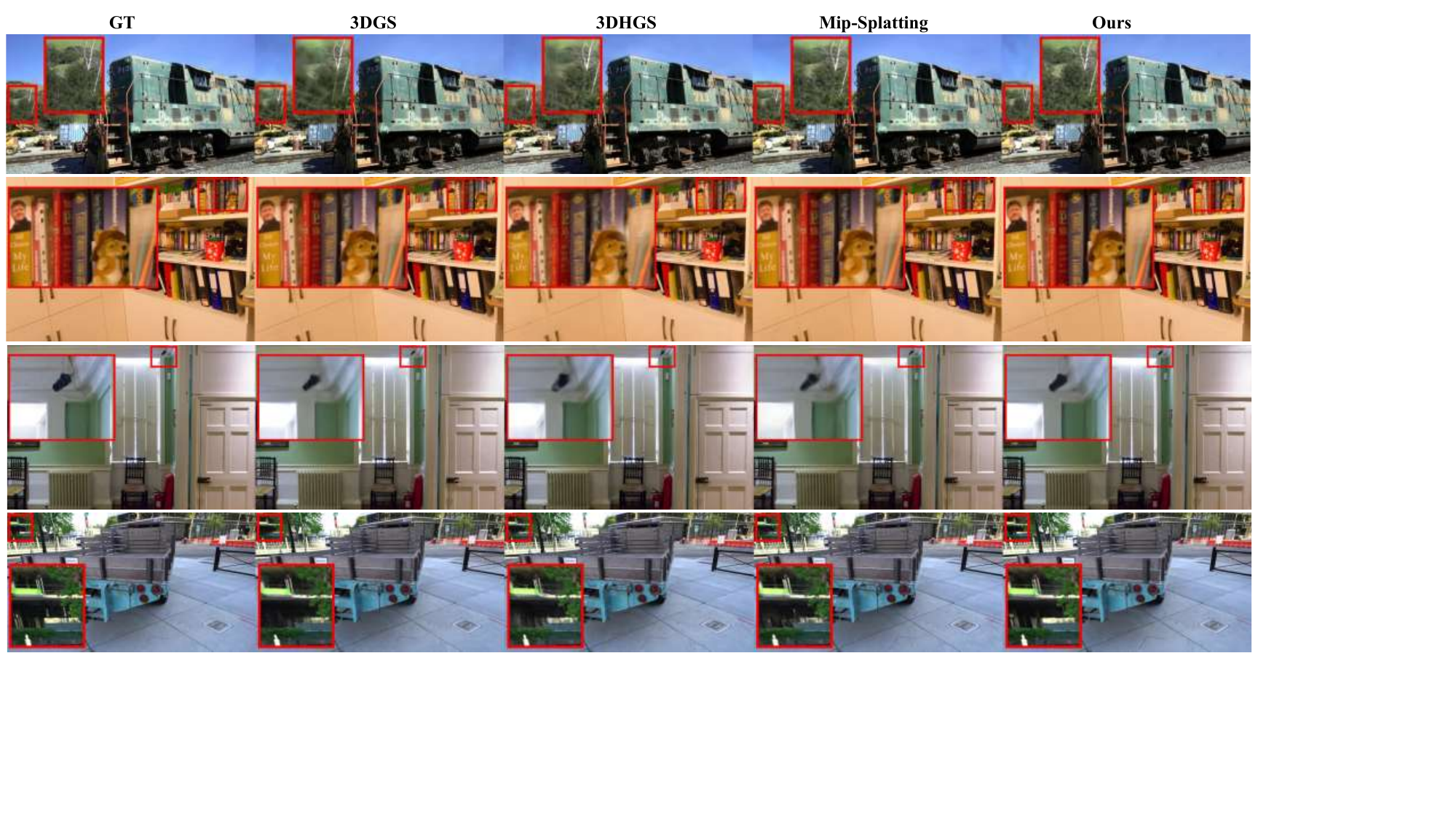}}
  \caption{\textbf{Comparison Results.} Visual differences are highlighted with red insets for better clarity. Our approach consistently outperforms other models on Tanks $\&$ Temples~\cite{Knapitsch2017} and Deep Blending~\cite{hedman2018deep} dataset, demonstrating advantages in challenging scenarios such as thin geometries and fine-scale details. Best viewed in color.}
\label{e2}
\end{figure*}

\end{document}